\documentclass[pdflatex,sn-mathphys-num]{sn-jnl}

\usepackage[inline]{enumitem}
\usepackage{graphicx}%
\usepackage{multirow}%
\usepackage{amsmath,amssymb,amsfonts}%
\usepackage{amsthm}%
\usepackage{mathrsfs}%
\usepackage[title]{appendix}%
\usepackage{xcolor}%
\usepackage{textcomp}%
\usepackage{manyfoot}%
\usepackage{booktabs}%
\usepackage{algorithm}%
\usepackage{algorithmicx}%
\usepackage{algpseudocode}%
\usepackage{listings}%
\usepackage{mathtools}
\usepackage{siunitx}
\usepackage{bm}
\newcommand{\boldlambda}{\boldsymbol{\lambda}}
\newcommand{\R}{\mathbb{R}}

\newcommand{\given}{\, | \,}
\newcommand{\trans}{^\mathrm{T}}
\newcommand{\diff}{\mathrm{d}}
\newcommand{\cupdot}{\mathbin{\mathaccent\cdot\cup}}

\newcommand{\bI}{\mathbf{I}}
\newcommand{\bk}{\mathbf{k}}
\newcommand{\bK}{\mathbf{K}}
\newcommand{\bx}{\mathbf{x}}
\newcommand{\bX}{\mathbf{X}}
\newcommand{\by}{\mathbf{y}}
\newcommand{\bz}{\mathbf{z}}
\newcommand{\cF}{\mathcal{F}}
\newcommand{\cT}{\mathcal{T}}
\newcommand{\E}{\mathbb{E}}
\DeclareMathOperator*{\expectation}{\mathbb{E}}
\newcommand{\Var}{\mathbb{V}}
\newcommand{\Prob}{\mathbb{P}}

\newcommand{\argmin}{\operatornamewithlimits{argmin}}
\newcommand{\PIMSE}{\(\text{PI}_{\text{MSE}}\)}

\newcommand{\ESTPDnorm}{EST-PD(Norm)\ }
\newcommand{\ESTPDhyp}{EST-PD(Hypsec)\ }
\newcommand{\revtext}[1]{\begingroup\color{black}#1\endgroup}
\newif\ifuncertainty
\newif\ifexclude
\newcommand{\newrevtext}[1]{\begingroup\color{black}#1\endgroup}

\theoremstyle{thmstyleone}%
\newtheorem{theorem}{Theorem}
\newtheorem{proposition}{Proposition}

\theoremstyle{thmstyletwo}%

\theoremstyle{thmstylethree}%
\newtheorem{definition}{Definition}%

\begin{document}

\title[Article Title]{Improving Random Forests by Smoothing}


\author*[1]{\fnm{Ziyi} \sur{Liu}}\email{ziyi.liu1@monash.edu}

\author[1]{\fnm{Phuc} \sur{Luong}}\email{felix.luong.cs@gmail.com}

\author*[2,1]{\fnm{Mario} \sur{Boley}}\email{mboley@is.haifa.ac.il}

\author*[1]{\fnm{Daniel F.} \sur{Schmidt}}\email{daniel.schmidt@monash.edu}

\affil*[1]{\orgdiv{Data Science and Artificial Intelligence}, \orgname{Faculty of Information Technology, Monash University}, \orgaddress{\street{Wellington Rd}, \city{Clayton}, \postcode{3800}, \state{Victoria}, \country{Australia}}}

\affil[2]{\orgdiv{Department of Information Systems}, \orgname{Faculty of Computer and Information Science, University of Haifa}, \orgaddress{\street{Abba Khoushy Ave.}, \city{Haifa}, \postcode{3103301}, \country{Israel}}}


\abstract{Random forest regression is a powerful non-parametric method that adapts to local data characteristics through data-driven partitioning, making it effective across diverse application domains. However, the piecewise constant nature of random forest predictions means each partition is predicted independently, ignoring potential smoothness in the underlying function. Particularly in the small data regime, this lack of information sharing across the input space can lead to suboptimal performance. In this work, we propose a kernel-based smoothing mechanism that enhances random forests by introducing local regularity to their predictions while preserving their adaptive partitioning capabilities. Our approach applies kernel smoothing to the piecewise constant outputs of random forests, effectively combining the adaptability of tree-based methods with the smoothness assumptions of kernel methods. We show that this smoothing procedure can be interpreted as capturing the variability/uncertainty in the tree cut points under resampling of the training inputs. Empirical results demonstrate that the proposed smoothed random forest model consistently improves predictive performance across diverse test cases, particularly in data-scarce settings.
Code, datasets, and experiment results are publicly available at \url{https://github.com/Neal-Liu-Ziyi/SmoothedRandomForest.git}.}




\maketitle

\section{Introduction}\label{sec1}

\newrevtext{Classical non-parametric regression models, such as Gaussian process~\cite{williams1995gaussian} and random forest regression~\cite{breiman2001random}, provide a flexible and robust approach to model response functions of unknown form that---in contrast to deep learning---tend to work well even if training data is limited~\citep{grinsztajn2022treebased}.
Non-parametric models can be characterised by their degree of spatial adaptivity and their smoothing behaviour~\cite{friedman1991multivariate,sherman2011spatial,chen2005adaptive}. In many applications, the behaviour of the target function varies substantially across the input space: some regions may be relatively flat, while others exhibit sharp peaks, valleys, or rapid local oscillations (see Figure~\ref{fig:smoothness_adaptivity}). Spatial adaptivity refers to the ability of a statistical model to adjust to such local variation in the data, thereby capturing heterogeneous smoothness and other local features~\cite{rovckova2023ideal,donoho1994ideal,loader2006local,lee2013locally}. This ability allows a model to preserve high-frequency structure where needed while reducing noise and limiting overfitting in smoother regions. Spatial adaptivity is therefore a key requirement for robust performance over broad function classes.
\begin{figure}[b!]
\centering
\includegraphics[width=0.95\textwidth]{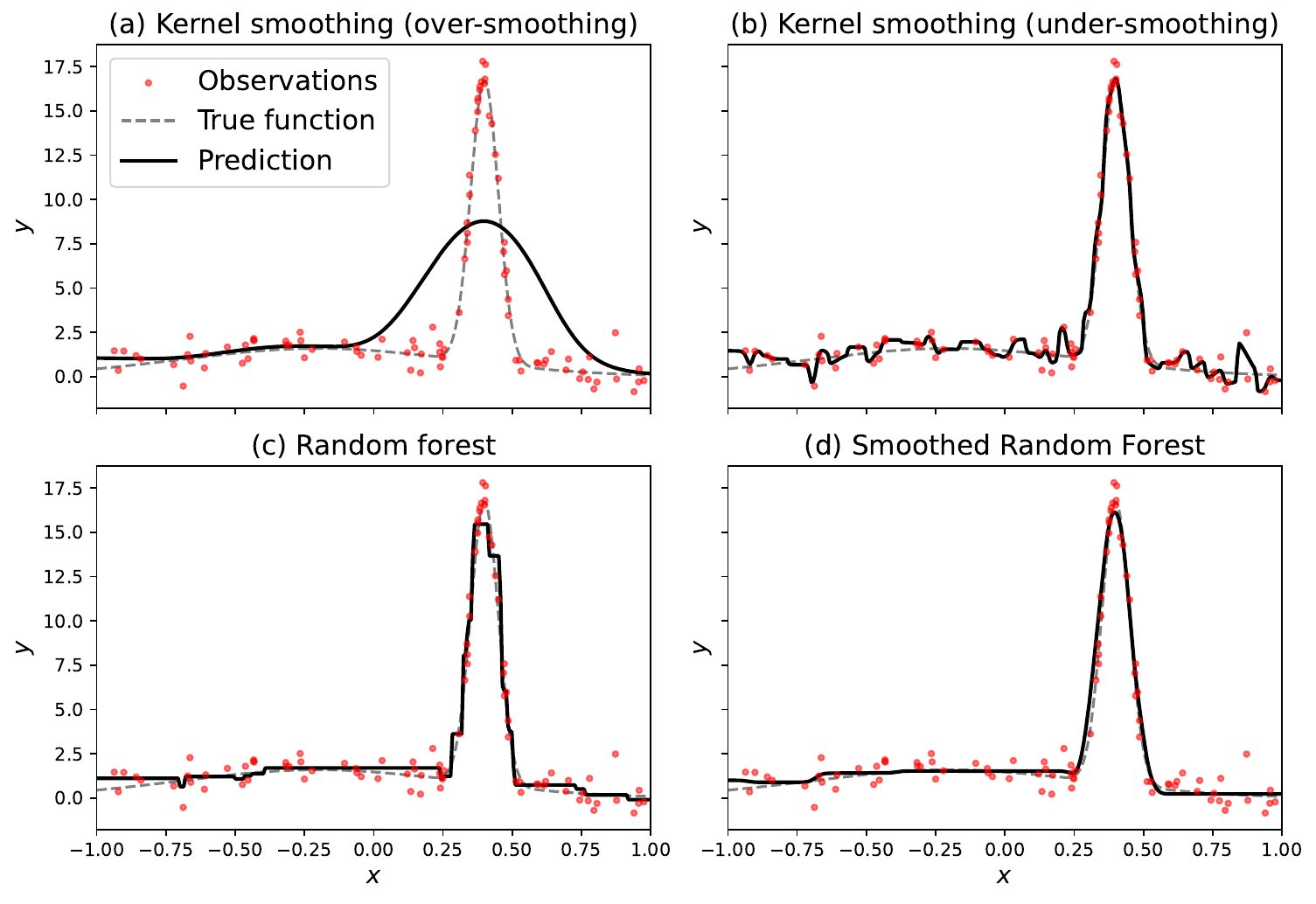}
\caption{\newrevtext{Trade-off between smoothness and spatial adaptivity: (a) over-smoothing fails to capture the sharp local peak, (b) under-smoothing overfits noise, (c) random forest produces discontinuities, (d) the proposed smoothed random forest model achieves both spatial adaptivity and smoothness.}}
\label{fig:smoothness_adaptivity}
\end{figure}

A range of methodological strategies has been developed to achieve spatial adaptivity. Penalised least-squares methods, such as variable-knot splines~\cite{mammen1997locally} and adaptive piecewise polynomial methods such as trend filtering~\cite{tibshirani_adaptive_2014}, allow the degree of local adaptation to respond to the structure of the signal, but do not easily scale to high dimensional input spaces. Classical smoothing methods achieve adaptivity by varying the effective smoothing scale across the domain, for example, through variable-bandwidth kernel estimators~\cite{lepski1997optimal} and data-driven local polynomial regression~\cite{fan1995data}. Gaussian processes with stationary kernels fall broadly into this class: they impose a single global smoothness regime and therefore tend to lack spatial adaptivity~\cite{pati2015optimal,rovckova2023ideal}. More adaptive behaviour can be introduced through nonstationary or input-dependent smoothing mechanisms~\cite{heinonen2016non}, but typically at the cost of increased modelling and computational complexity~\cite{szabo2025adaptation}.

Tree-based methods, including random forest as well as a range of gradient boosting methods~\citep[e.g.,][]{chen2016xgboost, ke2017lightgbm, prokhorenkova2018catboost}, provide a simpler route to spatial adaptivity through recursive partitioning of the input space and the fitting of simple models within each region. 
This approach enables spatial adaptivity by concentrating modelling effort in relevant regions of the predictor space~\cite{scott2004adaptive}.
Random forests are a particularly popular tree-based method due to their robustness, computational efficiency, and analytical tractability as adaptive local averaging method~\cite{lin2006random,scornet2016random}. 
However, similar to all other classical tree-based methods, random forests produce piecewise-constant prediction functions with discontinuities at partition boundaries.
This lack of smoothness, which is emphasised in the small data regime, can hinder the effective modelling of gradual transitions and continuous response surfaces~\cite{friedberg2020local}. 

To address this issue, multiple smooth tree variants have been proposed including trees based on soft or fuzzy splits, smooth transition mechanisms, and probabilistic assignment of observations to multiple regions~\cite{irsoy2012soft,olaru2003complete,da2008tree,alkhoury2020smooth,linero2018bayesian}. By replacing hard partition boundaries with continuous gating or weighting functions, these methods produce smoother predictive surfaces while retaining some of the local adaptivity of trees. However, by departing from standard recursive partitioning, these methods can no longer rely on the highly efficient, robust, and well-understood tree-growing algorithms used in CART, random forests, and related ensemble methods. Instead they have to resort to more complicated fitting schemes that are more computationally demanding and less stable than conventional tree-based methods.


In contrast, this article introduces a post-learning smoothing approach for random forest predictions. Rather than altering the construction of the forest itself, the proposed approach applies kernel-based smoothing to the piecewise-constant predictor learned by the forest. Like soft-trees, it achieves smoothness and spatial adaptivity, and is suitable for the small data regime. However, in contrast to  soft-trees, it does so while directly leveraging the efficiency, simplicity and robustness of random forests.
%
%
In more detail, this article:
\begin{enumerate}
    \item introduces probabilistic kernel smoothing as general approach to equip any learned prediction function with an arbitrary degree of smoothness while being efficiently computable, in particular, for piecewise constant functions,
    \item analyses how this smoothing procedure can be viewed as capturing the variability/uncertainty in tree cut points under resampling of the training inputs,
    \item investigates the out-of-bag performance of random forest ensemble members as effective criterion to select the degree of smoothness, and
    \item evaluates the approach on a range of benchmark dataset, demonstrating that probabilistic kernel smoothing tends to substantially improve the predictive performance of random forests and to do so more effectively than increasing the number of trees (by a factor of ten).
\end{enumerate}
}

%
%

\newrevtext{\section{Smoothing Function Estimates}\label{sec3}}


\newrevtext{In this section, we introduce the proposed smoothing framework on a general level before applying it to random forests in Sec.~\ref{sec4}. 
After recalling some basics about non-parametric function estimation (Sec.~\ref{sec:preliminaries}), we define 
the core smoothing mechanism (Sec.~\ref{sec:smoothing_general}) followed by its implementation
for piecewise constant functions such as decision trees (Sec.~\ref{subsec3.2}).
Finally, we provide some theoretical motivation and discuss the choice of smoothing kernel (Section~\ref{subsec3.3}).}

\begin{figure}[t]
\centering
\includegraphics[width=1\textwidth]{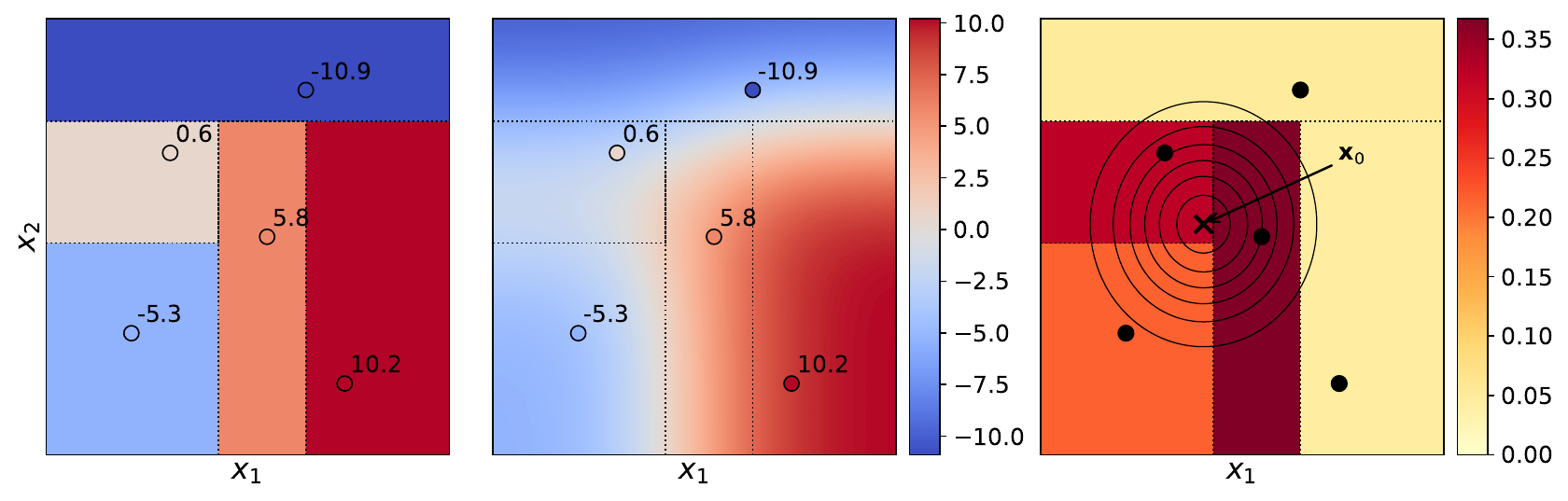}
\vspace{-0.4cm}
\caption{Illustration of smoothing mechanism applied to decision tree regressor: original piecewise constant predictions (left), smoothed and recalibrated predictions (middle) and leaf probabilities for specific query point $x_0$ (right). Training points (circles) are annotated with corresponding $y$-values.}
\label{1d_example}
\end{figure}
\newrevtext{\subsection{Non-parametric Noisy Function Estimation}\label{sec:preliminaries}}


\newrevtext{We consider a standard regression setting where one is 
interested in pointwise estimation of the regression function 
$g(\bx) = \E[y \given \bx]$, defined by the conditional distribution 
of a continuous random variable $y$ given a $p$-dimensional input vector $\bx$.
Specifically, we consider non-parametric estimators or prediction functions $f(\bx \given \bX, \by)$ fitted through a training dataset 
$(\bX, \by)$ of $n$ independent realisations of the joint distribution 
of $\bx$ and $y$. One important example of such an estimator is given by Gaussian process regression, defined as
\begin{equation*}
f_\mathrm{GPR}(\bx_0 \given \bX, \by) = \bk_0\trans (\bK + \sigma^2\bI)^{-1}\by
\end{equation*}
based on a noise variance parameter $\sigma^2$ and a covariance kernel $k(\bx, \bx')$ that gives rise to a kernel matrix $\bK$ with entries $k_{i,j}=k(\bx_i, \bx_j)$ and similarity vector $\bk_0=(k(\bx_0, \bx_1), \dots, k(\bx_0, \bx_n))\trans$.
This estimator represents a global kernel-smoothing technique that defines $f$ as a similarity-weighted linear combination of (de-correlated) outputs $(\bK + \sigma^2\bI)^{-1}\by$.
Crucially, the typically used covariance kernels are stationary, i.e., invariant to translations, which implies a globally uniform variability of the estimator $f$.

\newcommand{\with}{\!:\,}
\newcommand{\bw}{\mathbf{w}}
\newcommand{\bom}{\mathbf{m}}
\newcommand{\boM}{\mathbf{M}}
In scenarios where this behaviour is undesired, random forest regression provides an alternative approach with spatially adaptive variability. 
The random forest estimator is typically defined as the average of regression tree estimators that are randomly fitted to bootstrap samples of the training data:
\begin{align}
f_\mathrm{RF}(\bx_0 \given \cF, \bX, \by) &= \frac{1}{T} \sum_{t=1}^T f(\bx_0 \given \cT_t, \bom_t, \bX, \by)\label{eq:rf}\\
f(\bx_0 \given \cT, \bom, \bX, \by) &= \sum_{i=1}^n \left( \frac{m_{i}\mathbb{I}(\cT(\bx_i)=\cT(\bx_0))}{\sum_{i=1}^m m_i\mathbb{I}(\cT(\bx_i)=\cT(\bx_0))} \right) y_i \nonumber\enspace .
\end{align}
Here $\bom_{t}$ denotes the count-vector where component $m_{i,t}$ indicates the number of times data point $i$ is selected in bootstrap sample $t$, $\cF=\{\cT_1,\dots, \cT_t\}$ denotes the set of $t$ randomly fitted tree structures corresponding to hierarchical partitions of the input space, and $\cT(\bx)$ denotes the region of partition $\cT$ in which input point $\bx$ resides.
Similar to kernel methods~\citep{scornet2016random}, the random forest estimator can also be written as a similarity-weighted linear combination of outputs $f_\mathrm{RF}(\bx_0)=\bw(\bx_0)\trans\by$ where the $i$-th component of the similarity vector $\bw(\bx_0)$ measures the proportion of trees that place $\bx_0$ and $\bx_i$ in the same leaf, or formally
\begin{equation*}
    w_i(\bx_0) = \frac{1}{T} \sum_{t=1}^T \frac{m_{t,i}\mathbb{I}(\cT(\bx_i) = \cT_t(\bx_0))}{\sum_{i=1}^n m_{t,j}\mathbb{I}(\cT(\bx_j) = \cT_t(\bx_0))} \enspace .
\end{equation*}
This form of smoothing achieves spatial adaptivity, but produces discontinuous, piecewise constant predictions---with larger jumps for coarser tree partitions, which typically arise in the small data regime.}

\newrevtext{
\subsection{Probabilistic Kernel Smoothing}
\label{sec:smoothing_general}
Such large discontinuities are often unrealistic for real-world regression functions and, hence, tend to be disadvantageous for an estimator.
For such scenarios, we consider in this work the convolution of any learned estimator $f$ with an appropriately chosen probability kernel as a general smoothing technique to create a differentiable or at least continuous prediction function.
In contrast to directly applying kernel smoothing on the raw training data, transforming an already learned prediction function allows to retain its beneficial features such as spatially adaptive behaviour.

In more detail, we consider probability kernels $k(\cdot \given \bx, \boldlambda)$ that define a location/scale family of distributions with location parameter $\bx$ and positive scale parameter $\boldlambda$. 
That is, the kernel $k(\cdot \given \bx, \boldlambda)$ is assumed to satisfy: (1) non-negativity across the entire domain; 
(2) $\int k(\bz \given \bx, \boldlambda)\, \diff\bz = 1$, 
i.e., it must be a valid probability density over $\R^p$;
and (3) 
\begin{equation*}
    k(\bz \given \bx, \boldlambda)=\prod_{j=1}^p \lambda^{-1}_i k(\mathrm{diag}(\boldlambda)^{-1}(\bz - \bx) \given \boldsymbol{0}, \boldsymbol{1}).
\end{equation*}
Two concrete kernels that are used below, are the Gaussian and the hyperbolic second kernels defined as
\begin{align*}
k_\mathrm{gs}(\mathbf{z} \given \bx, \boldsymbol{\lambda}) &= \prod_{j=1}^p \left(\frac{1}{{2\pi\lambda_j^2}} \right)^{\frac{1}{2}} \exp\left(-\frac{(z^{(j)} - x^{(j)})^2}{2\lambda_j^2}\right) \\
k_\mathrm{hs}(\mathbf{z} \given \bx, \boldsymbol{\lambda}) &= \prod_{j=1}^p \left( \frac{1}{2\lambda_j} \right) \operatorname{sech}\left( \frac{\pi (z^{(j)} - x^{(j)})}{2\lambda_j} \right) \enspace .
\end{align*}
These two examples are product kernels that can be written as product of independent one-dimensional kernels, which will be computationally convenient in what follows.
However, the following central concept of a smoothing transformation is defined for probability kernels more generally.

\begin{definition}[\textbf{Probabilistic Smoothing}]
\label{def:smoothed:prediction}
Given a prediction function $f(\cdot)$ and the probability kernel $k(\cdot \given \bx, \boldlambda)$ of a location/scale family, the \textbf{probabilistically smoothed prediction} 
$\tilde{f}(\bx_0 \given \bX, \by, \boldlambda)$ is defined as the expectation of $f(\bz)$ for a random input $\bz$ with distribution given by the kernel 
at location $\bx_0$ with scale $\boldlambda$, i.e.,
\begin{align}
\tilde{f}(\bx_0\given \bX,\by,\boldsymbol{\lambda})
= \expectation_{\bz\sim k(\cdot \given \bx_0, \boldlambda)}\left[f(\bz\given \bx_0,\bX,\by)\right]
= \intop_{\R^p} f(\bz\given\bX,\by) k(\bz\given\bx_0,\boldsymbol{\lambda})\diff \bz. 
\label{general_form} \enspace .
\end{align}
\end{definition}

As an illustration of the definition, Fig.~\ref{1d_example} depicts the smoothing process when applied to a two-dimensional decision tree regressor.
Note that the scale $\boldlambda$ is a free parameter of the transformation.
In the limit for $\|\boldlambda\| \to 0$, the smoothed prediction recovers the original prediction.
Conversely, for $\|\boldlambda\| \to \infty$, the smoothed prediction becomes the identical global average prediction for all input points.
Hence, we can also think of $\boldlambda$ as \textbf{smoothing parameter}.

It is important to note that, while the probabilistic smoothing transformation yields smooth prediction functions that inherit differentiability from the employed probability kernel (Sec.~\ref{sec:differentiability:smoothed}), it can shrink the output of the original prediction function and also change its overall mean value.
Indeed, for square integrable\footnote{While tree models are generally not square integrable, one can restrict their support to establish the $L_2$-shrinkage property (Sec.~\ref{sec:basic_properties}).} $f \in L_2$ and symmetric kernels we have that $\|\tilde{f}\|_2 \leq \|f\|_2$ (Sec.~\ref{sec:basic_properties}).
For that reason, we will consider \textbf{re-calibrated smoothed predictions} in the reminder of this work, 
\begin{align}
\hat{y}(\bx_0\given \bX,\by,\boldsymbol{\lambda}, \beta_0,\beta_1)
&= \beta_1\tilde{f}(\bx_0\given \bX,\by,\boldsymbol{\lambda}) + \beta_0. 
\label{general_calibrated_pred} \enspace ,
\end{align}
i.e., affine linear transformations of $\tilde{f}$ with parameters $\beta_0 \in \mathbb{R}$ and $\beta_1 \in \mathbb{R}_+$.
}

\subsection{Piecewise Constant Prediction Functions}\label{subsec3.2}
We now specialise Equation~\eqref{general_form} to the case of piecewise 
constant functions. \newrevtext{A function is called \textbf{piecewise constant} 
if its domain $D \subseteq \R$ can be decomposed into $k \in \mathbb{N}$ disjoint but individually contiguous
regions $D=D_1 \cupdot \cdots \cupdot D_k$ with corresponding constants 
$c_1,\cdots c_k$ such that $f(\bx)=c_i$ if $\bx \in D_i$~\cite{tai2003multiple,
tamanini1996optimal}.
In one dimension, contiguous regions correspond to intervals, and piecewise constant functions therefore are step functions with abrupt changes or discontinuities at the boundaries of these intervals.

For higher dimensional inputs, contiguous regions can have arbitrarily complicated shapes. However, here we focus on regions that are \textbf{axis-parallel hyper-rectangles}, i.e., those given as Cartesian products of intervals
\begin{equation*}
D_i = \bigtimes_{j=1}^p [l^{(j)}_i, u^{(j)}_i] \enspace .
\end{equation*}
where the lower and upper bounds, $l^{(j)}_i, u^{(j)}_i$, are taken from the extended real line $\R \cup \{-\infty, \infty\}$.
This restriction covers the focus of this work, decision trees and forests, and allows us to tackle them in a computationally convenient way.
However, the discussed approach can be generalised to more complicated shapes.
}
%
%

\newrevtext{If the prediction function is piecewise constant, the smoothed prediction function~\eqref{general_form} can be rewritten as a weighted sum over each region, where the weight of each region \(D_i\) is the probability, \(\Prob(\bz\in D_i \given \bx_0,\boldlambda)\), that the latent variable \(\bz\), which follows a distribution centred at query point \(\bx_0\) and scaled by smoothing parameter \(\boldlambda\), falls into the region, i.e.,
\begin{align}
    \tilde{f}(\bf{x}_0\given \bf{X},\bf{y},\boldsymbol{\lambda}) 
\label{piece_prediction}
            &=\sum_{i=1}^k c_i \ \Prob({\bf z}\in D_i \given \bx_0,\boldsymbol{\lambda}).
\end{align}
This simplification means that as long as the probabilities of the latent random variable $\bz$ falling into the individual regions can be efficiently computed, so can be the smoothed predictions. 
In particular, this is the case for product kernels $k(\bz \given \bx, \boldlambda)=\prod_{j=1}^p k_j(z_j \given x_j, \lambda_j)$. In this case, the probabilities can be computed as
\begin{equation}
    \label{eq:prob:norm:kernel}
    \Prob({\bf z}\in D_i \given \bx_0,\boldlambda) =\prod_{j=1}^{p} \left[ K_j(u_i^{(j)} \given \bx_0^{(j)}, \lambda_j) - K_j(l_i^{(j)} \given \bx_0^{(j)}, \lambda_j) \right],
\end{equation}
where $K_j(u \given x, \lambda_j)=\int_{-\infty}^u k(z \given x, \lambda_j)\diff z$ denotes the cumulative distribution function of the $j$-th kernel factor. 

Using \eqref{piece_prediction} and \eqref{eq:prob:norm:kernel}, the time complexity for producing a prediction at a query point $\bx_0$ is $O(kp)$, where $k$ is the number of regions in the piecewise constant function. This is in contrast to other smoothing techniques, such as GPs, which require $O(n^3)$ operations per prediction. However, compared with random forests, which require, on average, only $O(\log n)$, the proposed model is less efficient for large data sets.}

\newrevtext{\subsection{Smoothing and Tree Structure Variability}\label{subsec3.3}}
\newcommand{\with}{\!:\,}
\newrevtext{The following analysis provides theoretical motivation for the 
kernel smoothing approach by showing that the smoothed prediction can be 
interpreted as capturing the variability in tree structure under resampling 
of the training data. We do this by examining the distribution of the break-point estimate of a one-cut decision tree under resampling from the following simple sampling distribution with univariate $x$ and deterministic $y \given x$ that is parameterised by a break-point $b$ and radius $w$:
\begin{align}
    x &\sim \mathrm{Uniform}(b-w, b+w) \nonumber\\
    y \given x &= g(x) = \mathbb{I}(x - b > 0) \enspace . \label{eq:sampling_thm1}
\end{align}
A one-cut decision tree (stump) estimator for $g(x)$ fitted to a training sample $\bX=(x_1, \dots, x_n)$, $\by=(g(x_1), \dots, g(x_n))$ then takes the form
\begin{align}
    {f}(x_0 \given \bX, \by)) &= \mathbb{I}(x_0 - \hat{b}(\bX, \by)>0)\nonumber\\
    \hat{b}(\bX, \by) &= (x_{\sup}(\bX,\by)+x_{\inf}(\bX,\by))/2\label{eq:estimate_thm1}
\end{align}
where the break-point estimate $\hat{b}$ defined as the mid-point between the largest $x$-sample less than $b$ and the smallest $x$-sample greater than $b$, i.e.,
\begin{align*}
x_{\sup}(\bX, \by)&=\sup \{x_i \with y_i = 0, 1 \leq i \leq n\}=\sup \{x_i \with x_i < b, 1 \leq i \leq n\} \enspace ,\\
x_{\inf}(\bX,\by)&=\inf \{x_i \with y_i = 1, 1 \leq i \leq n\}=\inf \{x_i \with x_i \geq b, 1 \leq i \leq n\} \enspace .
\end{align*}
%
\begin{figure}[t!]
\centering
\includegraphics[width=1\columnwidth]{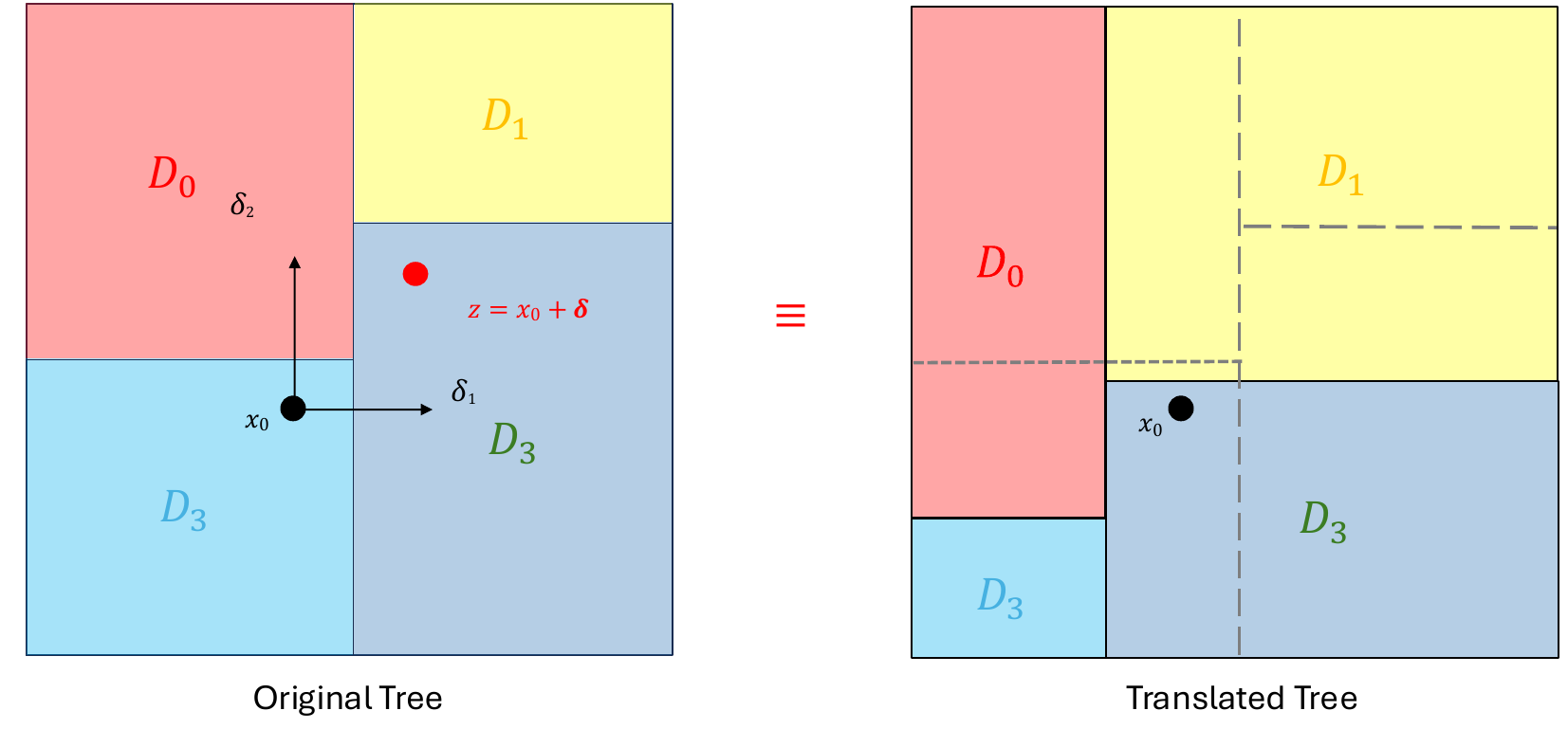}
\caption{\newrevtext{Illustration of the equivalence between evaluating the original 
piecewise function $f(\cdot)$ at a perturbed query point 
$\bz = \bx_0 + \bm{\delta}$ (left) versus evaluating the translated 
function $\bar{f}(\cdot)$ with shifted boundaries at the original query 
point $\bx_0$ (right). Both approaches yield identical predictions, 
demonstrating how kernel smoothing with latent variable $\bz$ mimics split point uncertainty.}}
\label{fig:translation:property}
\end{figure}
%
Though it is very simple, this model captures the local behaviour of a decision tree, in the sense that splitting a leaf in the typical top-down tree induction process is equivalent to fitting a decision stump to the reduced set of data in the leaf. A similar stump model was used in previous work analysing split points~\citep{banerjee2007confidence}; however, the following result, while less general, is more accessible, as the sampling distribution derived in~\citet{banerjee2007confidence} is only implicitly defined through a maximisation problem over a Brownian motion process.
Specifically, we have the following result regarding the asymptotic distribution of $\hat{b}$ (proven in Sec.~\ref{sec:proof_thm1}). \\
\begin{theorem}\label{theorem:1}
Let $\hat{b}=\hat{b}(\bX, \by)$ be the break-point estimate defined in Eq.~\ref{eq:estimate_thm1} fitted to a sample $\bX$ of size $n$ from the uniform distribution on $(b-w, b+w)$ with deterministic labels $\by=(g(x_1),\dots, g(x_n))$ as defined in Eq.~\ref{eq:sampling_thm1}.
Then, 
\begin{equation*}
    n (\hat{b} - b) \overset{d}{\to} w \, L
\end{equation*}
as $n \to \infty$, 
where $L$ follows a Laplace distribution with location 0 and scale 1.
\end{theorem}
\noindent After this analysis of the variation of tree break-points, we now turn to see how such variation is captured in the proposed smoothing process using the latent variable $\bz$. For that, note that the latent variable $\bz = \bx_0 + \bm{\delta}$, where $\bm{\delta} \sim k(\cdot \given {\bf 0}, \boldlambda)$. If $f$ is a piecewise constant function, }evaluating it at a query point $\bz$ is equivalent to evaluating the {\em translated} piecewise function
\begin{equation}
    \label{eq:translated:tree}
    \bar{f}(\bx) = c_i \mbox{\; if \;} \bx \in \left\{ \bx' : \bx' - \bm{\delta} \in D_i \right\}
\end{equation}
at the original query point $\bx_0$. 
\newrevtext{This equivalence is 
illustrated in Figure~\ref{fig:translation:property}, which shows how 
evaluating the original tree at the perturbed point $\bz = \bx_0+\bm{\delta}$ 
(left panel) produces the same prediction as evaluating the translated tree 
with shifted boundaries at the original point $\bx_0$ (right panel).} 
Therefore, by choosing the distribution (i.e., the kernel) of $\bz$ 
appropriately, we can mimic the effect of resampling the design points 
has on the estimated break-points (leaf boundaries) of a tree.

Theorem~\ref{theorem:1} states that the limiting distribution for $n \to \infty$ of the estimate $\hat{b}(\bX, \by)$ is of Laplace form. Therefore, at least for large $n$, due to the translation property~\eqref{eq:translated:tree}, the expectation of the prediction made by our proposed model at point $\bx_0$ under resampling of the data (denoted by $\E_{{\bX},{\by}}$) is 
%
\begin{align*}
\E_{{\bX},{\by}} [ f(\bx_0; \hat{b})] \to \, \mathbb{E}_\delta[ f(\bx_0; b + \delta) ] =\, \mathbb{E}_\delta[ f(\bx_0-\delta; b) ].
\end{align*}
%
\noindent as $n \to \infty$, where $b$ is the population value of the split-point, $\delta$ follows a Laplace distribution as per Theorem~\ref{theorem:1}, and the equality follows from translation property~\eqref{eq:translated:tree}. Of course we do not know the population value of $b$, but from Theorem~\ref{theorem:1} we know that $\hat{b}$ is a consistent estimator of $b$, so we may replace $b$ by $\hat{b}$ and arrive at the following estimator for $\mathbb{E}_{{\bX},{\by}}[ f(\bx_0; \hat{b})]$ ,
\begin{equation*}
\begin{split}
&\hat{\mu}(\bx_0) = \mathbb{E}_\delta[ f(\bx_0 - \delta; \hat{b}) ]\ 
\end{split}
\end{equation*}
\noindent which is exactly equal to (\ref{piece_prediction}), if $\delta$ follows the Laplace distribution from Theorem~\ref{theorem:1}. 

\newrevtext{Theorem~\ref{theorem:1} suggests using a spherical Laplace distribution, at least asymptotically; however, the Laplace distribution exhibits a sharp peak that forms a non-smooth cusp at the origin, making it unsuitable for applications requiring differentiability. To ensure smoothness while preserving the desirable properties of the Laplace distribution, we instead consider two alternatives: spherical Gaussian and hyperbolic secant distributions. Both distributions are unimodal, bell-shaped, and everywhere differentiable, and both can adequately approximate the Laplace distribution over compact regions of the input space with appropriate scale selection.
Importantly, the hyperbolic secant distribution exhibits exponential tail decay, matching the asymptotic behaviour of the Laplace distribution suggested by Theorem~\ref{theorem:1}; in contrast, the Gaussian distribution has lighter tails. However, in contrast to the Laplace distribution, the hyperbolic secant kernel maintains infinite differentiability everywhere, including at the origin.}

\newrevtext{
\section{Application to Random Forests}\label{sec4}

This section applies the methodology in Section~\ref{sec3} to ensembles of trees, specifically, random forests. We first discuss how predictions of a smoothed ensemble model can be found efficiently as the averaged predictions of smoothed ensemble members. Then we discuss model selection specifically for the context of random forests.

\subsection{Smoothing Additive Ensembles}\label{subsec4.1}
As defined in \eqref{eq:rf}, random forests are additive ensembles of individual tree models.
As the average of piecewise constant functions, the random forest prediction function is itself a piecewise constant function.
However, the number of its regions can grow exponentially with the dimensionality of the data.
To see this, consider the case of an ensemble of one decision stump per dimension 
\begin{equation}
f_t(\bx)=c_{+,i}\mathbb{I}(x_t \geq l_t) + c_{-,i}\mathbb{I}(x_t < l_t)
\end{equation}
for $1 \leq t \leq p=T$. 
In this case, though each ensemble member has only two regions, the overall ensemble has $2^p$ regions corresponding to the combinations of possible outcomes of the checks $x_t \geq l_t$ for $1 \leq t \leq p$.
It is thus computationally prohibitive to apply smoothing via \eqref{piece_prediction} and \eqref{eq:prob:norm:kernel} directly to the ensemble regions.
Fortunately, the probabilistic smoothing transformation is a linear operator, as can be checked quite easily (see \ref{sec:proof_linearity_smoothing}):}
\newrevtext{\begin{proposition}\label{prop:global_local_relationship}
Let $f(\bx)=\sum_{t=1}^T w_t f_t(\bx)$ be an additive ensemble model with real coefficients $w_t \in \R$. Then $\tilde{f}(\bx \given \boldlambda)=\sum_{t=1}^T w_t \tilde{f}_t(\bx \given \boldlambda)$.
\end{proposition}}
\newrevtext{Thus, for any additive ensemble $f(\bx)=\sum_{t=1}^T w_t f_t(\bx)$ of piecewise constant functions $f_t$ we can find the smoothed ensemble prediction function by applying \eqref{piece_prediction} and \eqref{eq:prob:norm:kernel} to the individual ensemble members $f_t$ and forming their average.
With this approach, the computational complexity is reduced from $O(kn)$ to $O((k_1 + \dots + k_t)n)$ where $k$ is the total number of regions of the ensemble and $k_1, \dots, k_t$ are the number of regions of the individual ensemble members.}

\newcommand{\boldLambda}{\boldsymbol{\Lambda}}
\newrevtext{Interestingly, this approach then allows for even more flexibility: instead of globally smoothing the whole ensemble with a fixed smoothing parameter $\boldlambda$, one can form an ensemble of smoothed individual ensemble members, i.e.,
\begin{equation}
\label{est_prediction}
        \tilde{f}(\bx_0\given \boldLambda) 
        = \frac{1}{T}\sum_{t=1}^T w_t\tilde{f}_t(\bx_0 \given \boldlambda_t)) \enspace . 
\end{equation}
Here $\boldlambda_t$ denotes the $t$-th row of a $(T \times p)$-smoothing-parameter matrix $\boldLambda$.
The predictions for this smoothed model can then be computed with the computational complexity mentioned above via
\begin{equation}
\label{srf_prediction}
        \tilde{f}(\bx_0\given\boldLambda ) 
        = \frac{1}{T}\sum_{t=1}^T w_t \left(\sum_{i=1}^k c_i^{t}\ \Prob(\bz\in D_i^t \given \bx_0,\boldlambda_t)\right),
\end{equation}
where $c_i^t$ and $D_i^t$ for $1 \leq i \leq k_t$ are the $i$-th prediction value and $i$-th region of ensemble member $t$, respectively.

The flexibility of the smoothed model can be determined by imposing constraints on the smoothing parameter matrix, which can be formalised by requiring it to come from a specific set of matrices $\mathcal{S} \subseteq \R^{T \times p}$. Here, we investigate the following four options:
\begin{enumerate}
    \item \textbf{Isotropically smoothed ensemble of trees}, i.e., using the same smoothing parameter $\lambda$ for all dimensions and ensemble members, as captured in the constraint set $\mathcal{S}_\mathrm{STE} = \{\lambda\mathbf{1}_T\mathbf{1}_p\trans \with \lambda \in \R_+\}$ with one degree of freedom.
    \item \textbf{Per-dimension smoothed ensemble of trees}, i.e., using the same smoothing parameter vector $\boldlambda$ for all ensemble members, leading to constraint set $\mathcal{S}_\mathrm{STE-PD} = \{\mathbf{1}_T\boldlambda\trans \with \boldlambda \in \R^p_+\}$ with $p$ degrees of freedom.
    \item \textbf{Ensemble of isotropically smoothed trees}, i.e., using one smoothing parameter $\lambda_t$ per ensemble member applied uniformly to all dimensions, leading to constraint set $\mathcal{S}_\mathrm{EST} = \{\boldlambda\mathbf{1}_p\trans \with \boldlambda \in \R^T_+\}$ with $T$ degrees of freedom.
    \item \textbf{Ensemble of per-dimension smoothed trees}, i.e., using one smoothing parameter per ensemble member and dimension, leading to $\mathcal{S}_\mathrm{EST-PD}=\{\boldLambda \with \boldLambda \in \R_+^{T \times p} \}$ with $Tp$ degrees of freedom.
\end{enumerate}
}

\begin{figure}
    \centering
    \includegraphics[width=1.0\linewidth]{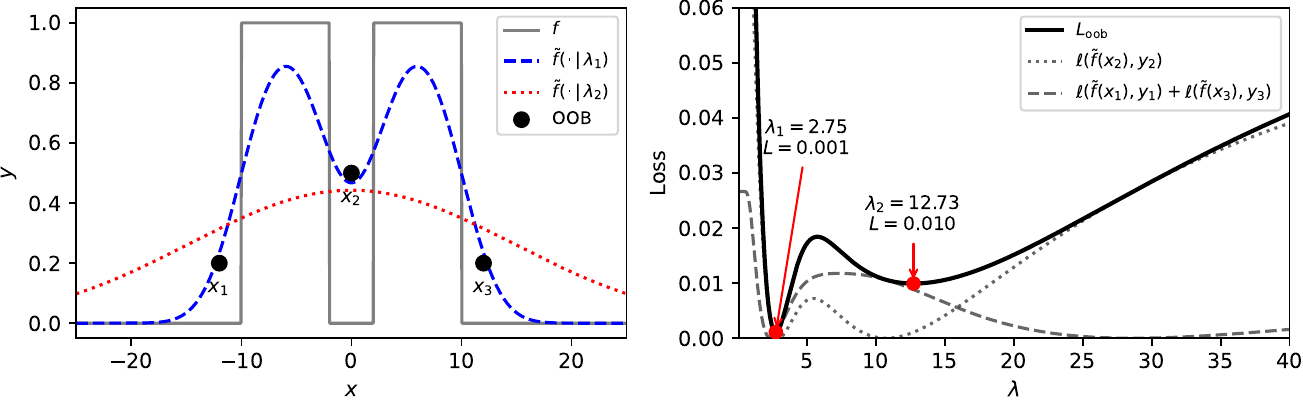}
    \caption{\newrevtext{Example of out-of-bag smoothing parameter optimisation with two local minima, $\lambda_1$ and $\lambda_2$, for 1-d inputs with Gaussian kernel smoothing.}}
    \label{fig:non_convexity}
\end{figure}
\newrevtext{\subsection{Model Selection}\label{sec:model_selection}

Depending on what smoothing constraints are chosen, it remains to select between $1$ and $Tp$ free parameters of the smoothing matrix $\boldLambda$ as well as the two re-calibration parameters $\beta_0$ and $\beta_1$.
While the latter can be chosen easily using the training data, doing the same for the former results in under-smoothing, because of the incentive to fit the training data as much as possible.

Fortunately, for random forests, we can use out-of-bag samples for individual trees; i.e., training example $i$ is out-of-bag for tree $t$ if it was not selected in the corresponding bootstrap sample, or formally, if $m_{i,t}=0$.
Optimising the out-of-bag performance per tree is a useful data-driven selection criterion because, in contrast to using all training data, the out-of-bag performance provides an unbiased estimate of the parameter values' generalisation performance.
Formally, denoting the indices of the pooled out-of-bag data as $\mathcal{O}=\{t, i \with m_{i,t}=0\}$, the out-of-bag loss is defined as
\begin{equation}
\label{eq:oob_risk}
L_\mathrm{oob}(\boldLambda, \beta_0, \beta_1) = \sum_{t,i\in\mathcal{O}}\left(y_{i} - \beta_1\tilde{f}(\bx_{i} \given \mathcal{T}_t,\bm{\lambda}_t) - \beta_0\right)^2 \enspace.
\end{equation}
While all parameters could be optimised jointly, we consider here a sequential process where we first optimise the smoothing parameters under the chosen model constraints $\mathcal{S}$, followed by the optimisation of the calibration parameters, i.e., to find
\begin{align}
\boldLambda^* &= \argmin \{L_\mathrm{OOB}(\boldLambda, 0, 1) \with \boldLambda \in \mathcal{S}\}\label{eq:select_smoothing}\\
\beta_0^*, \beta_1^* &= \argmin \{L_\mathrm{OOB}(\boldLambda^*, \beta_0, \beta_1) \with \beta_0, \beta_1 \in \R \times \R\} \enspace .\label{eq:select_calib}
\end{align}

Given the values of the smoothing parameters from \eqref{eq:select_smoothing}, the solution for the calibration parameters \eqref{eq:select_calib} are the simple least square regression parameters for the pooled out-of-bag data.
Defining the mean out-of-bag target values and predictions as $\bar{y}_\mathrm{oob}=\sum_{t, i \in \mathcal{O}} y_i/|\mathcal{O}|$ and $\bar{f}_\mathrm{oob}=\sum_{t,i \in \mathcal{O}}\tilde{f}_t(\bx_i,\given \boldLambda^*)/|\mathcal{O}|$, the solutions are:
\begin{align}
\beta_1^* &= \frac{\sum_{t,i \in \mathcal{O}} (\tilde{f}_{t}(\bx_i \given \boldLambda^*) - \bar{f}_\mathrm{oob})(y_i - \bar{y}_\mathrm{oob})}{\sum_{t,i \in \mathcal{O}}(\tilde{f}_{t}(\bx_i \given \boldLambda^*) - \bar{f}_\mathrm{oob})^2}\\
\beta_0^* &= \bar{y}_\mathrm{oob} - \beta_1^* \bar{f}_\mathrm{oob} \enspace .
\end{align}

In contrast, the optimisation of the smoothing parameters~\eqref{eq:select_smoothing} poses a complex non-convex optimisation problem. 
Indeed, even for 1-dimensional $f$ and correspondingly a single smoothing parameter, we can end up with multiple local minima of the out-of-bag loss (see Fig.~\ref{fig:non_convexity} for an instructive example).
On the other hand, the out-of-bag loss is differentiable in the smoothing parameter matrix.
Thus, as laid out in Appendix \ref{lambda_opt}, we can use heuristic gradient-based optimisation techniques to find or at least approximate $\boldLambda^*$.
Given the immense progress achieved in these methods in the context of deep learning, this approach allows to pick suitable optimisers from a wide range of options to achieve a suitable compromise between model selection accuracy and computational complexity.}

\section{Experiments}



\revtext{This section evaluates the EST-PD (ensemble of smoothed trees with per-dimension parameters) smoothing approach within the smoothed random forest framework. We focus on EST-PD, which demonstrates the most robust performance across diverse problem settings among the four smoothing strategies (see appendix~\ref{apdx:detaile_exp}). EST-PD is implemented using both Gaussian and hyperbolic secant kernels in 10-tree random forests. Performance is benchmarked against standard random forests with 10, 20, 50, and 100 trees, as well as Gaussian Process regression, across 10 UCI datasets. These benchmark ensemble sizes are chosen specifically to assess whether smoothed random forests with smaller ensembles can achieve competitive performance relative to standard random forests with substantially larger tree counts. Training set sizes range from 50 to 500 observations. Detailed experimental comparisons of all four smoothing strategies (STE, STE-PD, EST, EST-PD) are provided in Appendix~\ref{apdx:detaile_exp}.}

\subsection{Datasets, Metrics and Model Variants}

The experimental evaluation employs 10 datasets from the UCI machine learning repository, selected to provide coverage across diverse domains and varying characteristics. The datasets span power generation (CCPP), biological systems (Fertility), environmental modelling (Forest Fires), environmental toxicology (QSAR aquatic and fish toxicity), real estate valuation, financial analysis (Stock Portfolio Performance), food science (Wine Quality - red and white), and engineering design (Yacht Hydrodynamics). \revtext{All random forest models have their maximum tree depth optimised via 5-fold cross-validation to minimise mean squared error, to provide strong baselines for comparison. We set $\texttt{max\_features}=0.8$ to ensure a degree of feature randomness at each split. All other hyperparameters follow scikit-learn defaults.}

Model performance is assessed through mean squared error (MSE), providing a standard measure of prediction quality. To quantify improvements relative to baseline models, we employ the Percentage Improvement of MSE (\PIMSE), For two models $M_c$ and $M_b$, where $M_b$ is the baseline, the percentage improvement is calculated as
\begin{align*}
    \text{PI}_{\text{MSE}}(M_c,M_b) = \frac{\text{MSE}(M_b)-\text{MSE}(M_c)}{\text{MSE}(M_b)}\times 100\%.
\end{align*}
A higher \PIMSE indicates greater improvement, while zero indicates equivalence. \revtext{Additionally, we evaluate the median absolute deviation (MAD) and the maximum absolute error to provide additional measures of prediction performance. Statistical significance of performance differences in terms of ranks is assessed using the Nemenyi post-hoc test with critical difference analysis. Median absolute deviation and maximum absolute error results can be found in Appendix~\ref{apdx:detaile_exp}.}

\revtext{
The experimental framework employs a systematic naming convention encoding the relevant characteristics of each model configuration. All smoothed random forest variants are applied to the same underlying 10-tree random forest (RF(10)) structure to ensure fair comparison. RF($n$) denotes standard random forests with $n$ trees ($n = 10, 20, 50, 100$) trained on full training observations. GP represents Gaussian process regression with Matérn kernel ($\nu=2.5$) and white noise kernel. The length scale and noise variance are chosen via maximum marginal likelihood. EST-PD(Norm) and EST-PD(Hypsec) denote smoothed random forests using the ensemble of smoothed trees with per-dimension parameters approach, implemented with Gaussian (Norm) or hyperbolic secant (Hypsec) kernels, respectively. Alternative smoothing strategies are evaluated in Appendix~\ref{apdx:detaile_exp}.
}

\newrevtext{\subsection{Results}}

\revtext{
We tested the competing methods on 10 datasets. For each dataset, we varied the training sample sizes from 50 to 500 observations, with each training sample created by bootstrap sampling from the original sample; the remaining observations not included in each training sample were then used to compute test scores for each of the methods. For each dataset and sample size combination, this process was repeated 100 times.

Based on these experimental results \ESTPDnorm and \ESTPDhyp outperform the baseline RF(10) in 93.0\% and 94.0\% of cases, respectively, demonstrating consistent effectiveness across a range of problems. When benchmarked against larger random forests, both variants outperform RF(20) in 81\% of cases, RF(50) in 69\% of cases, and RF(100) in 64\% of cases, indicating that smoothing enables 10-tree ensembles to compete effectively with substantially larger forests, though the advantage diminishes with increasing ensemble size. Both EST-PD variants substantially outperform GP regression, achieving superior performance in 88\% of cases. Examining overall best performance across all methods, EST-PD variants collectively achieve the best results in 51.4\% of cases (2,723 out of 5,300), with EST-PD(Hypsec) winning 26.6\% and EST-PD(Norm) winning 24.8\%. RF(100) achieves the best performance in 23.5\% of cases, followed by GP (11.3\%), RF(50) (9.1\%), RF(20) (3.6\%), and RF(10) (1.1\%). The two kernel functions demonstrate remarkably similar performance profiles across all benchmarks, suggesting minimal impact of kernel choice on overall effectiveness.
}

\begin{figure}[ht]
\centering
\includegraphics[width=0.8\columnwidth]{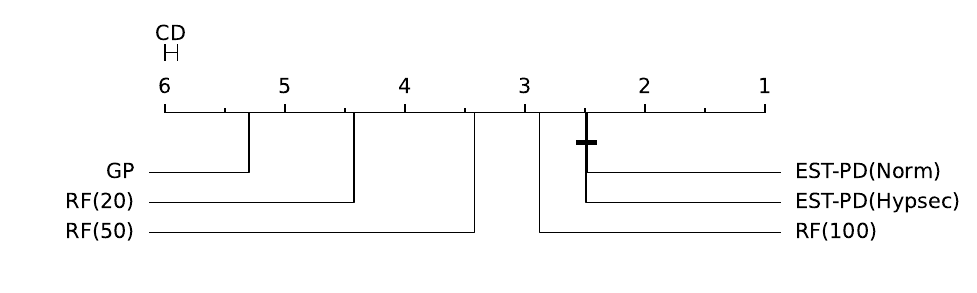}
\caption{\revtext{Critical difference diagram showing statistical comparison of methods using the Nemenyi post-hoc test ($\alpha=0.05$) based on percentage improvement in MSE relative to RF(10). Methods connected by horizontal bars are not statistically different. Numbers indicate mean rank across all experimental conditions (lower is better). The critical distance ${\rm CD} = 0.104$ represents the minimum rank difference required for statistical significance.}}
\label{fig:critical_difference}
\end{figure}

\revtext{
To provide statistical context for these comparisons, we conducted a nonparametric evaluation using the Friedman omnibus test followed by Nemenyi post-hoc pairwise comparisons (detailed results in Appendix~\ref{sec:statistical_analysis}). The analysis is based on \PIMSE \ relative to RF(10), enabling scale-free comparisons across datasets with different MSE ranges. Figure~\ref{fig:critical_difference} presents the critical difference diagram, showing mean ranks and statistical groupings with critical distance ${\rm CD}=0.104$ at $\alpha = 0.05$ significant level. EST-PD(Norm) achieves the best mean rank (2.482), followed by EST-PD(Hypsec) (2.493), RF(100) (2.879), RF(50) (3.420), RF(20) (4.424), and GP (5.302). The Nemenyi post-hoc test indicates that EST-PD(Norm) and EST-PD(Hypsec) form a statistical equivalence group seperate from the other methods. This statistical equivalence confirms that the two kernel variants exhibit no significant difference in performance. In terms of median \PIMSE~over RF(10) (over all datasets and samples sizes, see Appendix~\ref{sec:statistical_analysis}), EST-PD(Norm) and EST-PD(Hypsec) achieve a score of 9.96\% and 9.99\% respectively, exceeding RF(100) at 7.32\%, RF(50) at 6.63\%, and RF(20) at 4.24\%. The concordance between mean ranks and median improvements indicates that the EST-PD variants are able to deliver a consistent and meaningful improvement over standard random forests.
}

\begin{figure*}[ht]
\centering
\includegraphics[width=1.0\textwidth]{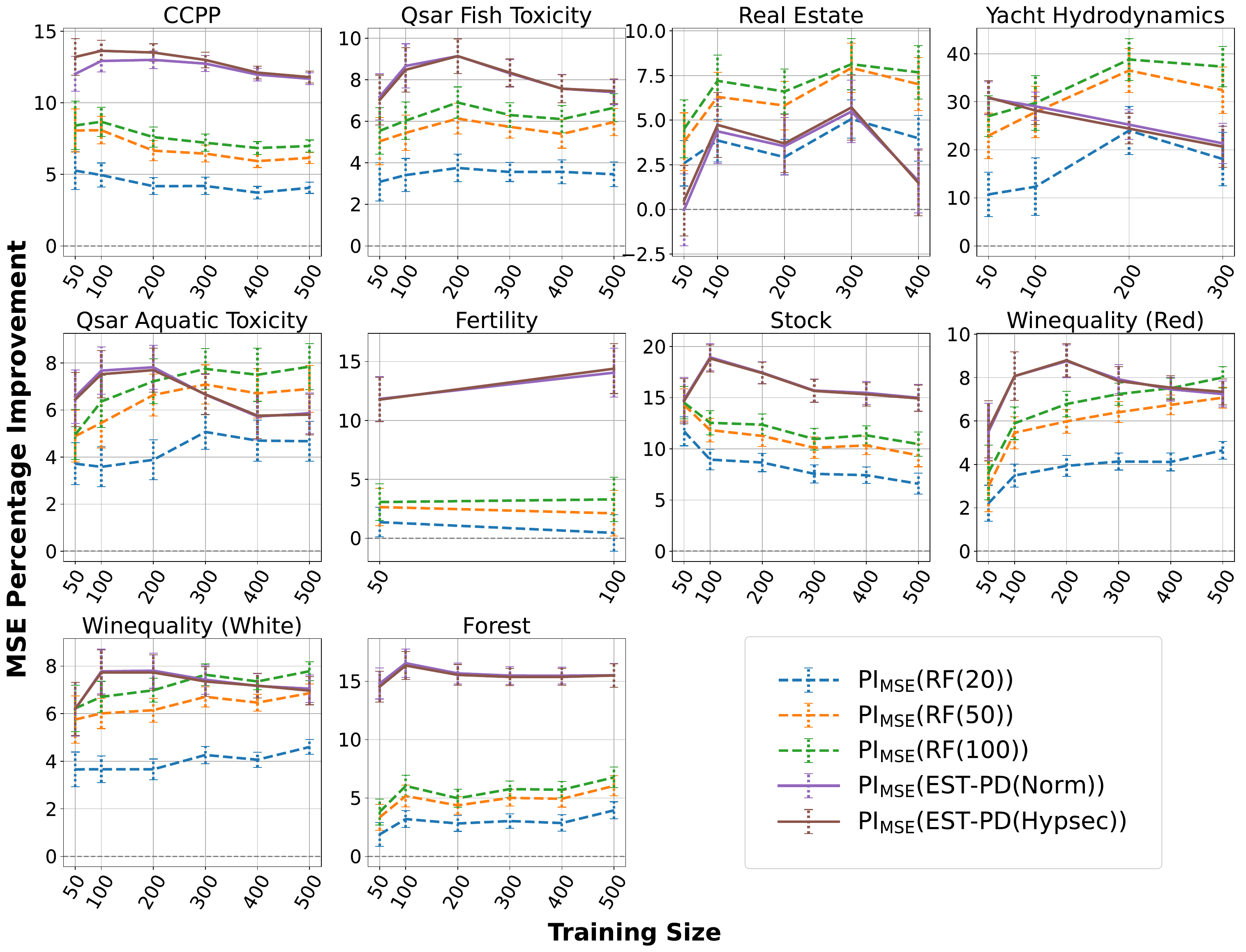} 
\caption{MSE percentage improvement over base model RF(10) for varying training sizes.}
\label{fig:PI_MSE_SE}
\end{figure*}

\revtext{
Table~\ref{tab:PI_MSE_SE} presents mean percentage improvements in MSE across different training sizes for each dataset, with standard errors reported in parentheses. Figure~\ref{fig:PI_MSE_SE} illustrates the performance trends as training size varies from 50 to 500 observations, focusing on random forest baselines and EST-PD variants. The GP baseline is omitted from the figure because it operates on a substantially different performance scale (often obtaining negative improvements indicating worse performance than the baseline), which would obscure distinctions among the more competitive methods. The results reveal some interesting dataset-specific heterogeneity in the effectiveness of smoothing, with performance patterns varying across problem domains.
}
%

\begin{table}[ht]
\resizebox{\columnwidth}{!}{%
\begin{tabular}{r c
    r@{.}l r@{.}l r@{.}l r@{.}l r@{.}l r@{.}l}
\hline
\multicolumn{1}{c}{Data} & P &
  \multicolumn{2}{c}{RF(20)} &
  \multicolumn{2}{c}{RF(50)} &
  \multicolumn{2}{c}{RF(100)} &
  \multicolumn{2}{c}{GP} &
  \multicolumn{2}{c}{EST-PD(Norm)} &
  \multicolumn{2}{c}{EST-PD(Hypsec)} \\ \hline
CCPP & 4
  & 4   & 40 (0.16) & 6   & 89 (0.18) & 7   & 62 (0.19)
  & $-76$   & 48 (4.10)  & 12  & 39 (0.15) & \textbf{12} & \textbf{88 (0.16)} \\
Qsar Fish Toxicity & 6
  & 3   & 47 (0.14) & 5   & 62 (0.17) & 6   & 25 (0.17)
  & $-19$   & 49 (0.73)  & \textbf{8} & \textbf{04 (0.18)} & 8 & 00 (0.18) \\
Real Estate & 6
  & 3   & 69 (0.27) & 6   & 17 (0.33) & \textbf{6} & \textbf{83 (0.34)}
  & $-100$  & 77 (1.75)  & 2   & 99 (0.42) & 3   & 22 (0.42) \\
Yacht Hydrodynamics & 6
  & 16  & 29 (1.38) & 29  & 94 (1.27) & \textbf{33} & \textbf{22 (1.21)}
  & $-284$  & 56 (14.84) & 26  & 60 (0.90) & 26  & 05 (0.91) \\
Qsar Aquatic Toxicity & 8
  & 4   & 27 (0.18) & 6   & 28 (0.21) & \textbf{6} & \textbf{94 (0.22)}
  & $-47$   & 49 (1.41)  & 6   & 72 (0.20) & 6   & 65 (0.20) \\
Fertility & 9
  & 0   & 91 (0.51) & 2   & 38 (0.64) & 3   & 19 (0.62)
  & $-1$    & 24 (1.34)  & 12  & 94 (0.72) & \textbf{13} & \textbf{08 (0.73)} \\
Stock & 11
  & 8   & 48 (0.22) & 11  & 20 (0.25) & 12  & 03 (0.25)
  & $-422$  & 89 (5.17)  & \textbf{16} & \textbf{22 (0.30)} & 16 & 14 (0.30) \\
Winequality (Red) & 11
  & 3   & 75 (0.11) & 5   & 78 (0.15) & 6   & 51 (0.16)
  & $-24$   & 66 (0.39)  & 7   & 49 (0.19) & \textbf{7} & \textbf{54 (0.19)} \\
Winequality (White) & 11
  & 3   & 98 (0.10) & 6   & 32 (0.12) & 7   & 11 (0.13)
  & $-17$   & 85 (0.36)  & \textbf{7} & \textbf{23 (0.16)} & 7 & 19 (0.16) \\
Forest & 12
  & 2   & 96 (0.16) & 4   & 81 (0.18) & 5   & 51 (0.18)
  & \textbf{18} & \textbf{98 (0.28)} & 15  & 59 (0.21) & 15  & 45 (0.21) \\ \hline
\end{tabular}%
}
\caption{Mean percentage improvement of MSE over the RF(10) base model 
across different training sizes. Standard errors are reported in 
parentheses. Bold indicates the best-performing method for each dataset.}
\label{tab:PI_MSE_SE}
\end{table}

The EST-PD variants demonstrate strong, stable performance across several datasets. On CCPP, both kernel variants maintain consistent improvements of 12-13\% across all training sizes, substantially exceeding RF(100) 7.6\% average improvement. Stock shows similar patterns with EST-PD improvements ranging from 15-19\%, peaking at smaller training sizes ($n=100$: 18.9\%) and gradually declining as sample size increases ($n=500$: 15.0\%), though remaining substantially above the RF(100) average of $12.0\%$. Forest exhibits remarkable stability, with EST-PD maintaining 15-16\% improvements consistently across all training sizes, significantly outperforming RF(100) with a 5.5\% average. Fertility, despite having only 100 total observations, shows particularly strong results with EST-PD achieving $11-14\%$ improvements compared to RF(100) 3.2\%, with performance increasing at larger training fractions.

In contrast, Real Estate presents a notably different pattern where smoothing provides inconsistent benefits. EST-PD improvements range from $-0.03\%$ at $n=50$ to $5.5\%$ at $n=300$, averaging only 3.0\%, compared to RF(100), a more stable $6.8\%$ average. This dataset represents the clearest case where additional trees in standard RF provide more reliable improvements than smoothing. The Wine Quality datasets show moderate but consistent EST-PD performance (Red: $7.5\%$, White: $7.2\%$) that matches or slightly exceeds RF(100) (Red: $6.5\%$, White: $7.1\%$).

Yacht Hydrodynamics exhibits an interesting training-size-dependence pattern. At small sample sizes, EST-PD achieves substantial improvements exceeding 30\%, surpassing RF(100) with 27\%. However, as training size increases, this advantage diminishes considerably: at $n=300$, EST-PD improvements decline to approximately 21\%. This suggests that smoothing provides particularly strong benefits in small-sample regimes but offers diminishing returns as more data become available for this dataset. QSAR datasets (Fish Toxicity, Aquatic Toxicity) show moderate EST-PD performance (8.0\% and 6.7\% respectively) that slightly exceeds or matches RF(100) performance.

\newrevtext{
\subsection{Discussion}

Overall, the experimental results strongly suggest that, in addition to producing smooth, differentiable prediction functions, post-hoc smoothing has the potential to improve the performance of random forests for a relatively low increase in computational overhead. Detailed MSE values, complete results across all training sizes, and additional performance metrics, including median absolute deviation and maximum absolute error, are provided in Appendix~\ref{apdx:detaile_exp}.

The two kernel variants (Gaussian and hyperbolic secant) demonstrate remarkably similar performance across all datasets and training sizes, with differences typically under $0.5$ percentage points. The largest observed difference is $0.9$ percentage points on Yacht Hydrodynamics at $n=50$, but such cases are rare. This consistency confirms that kernel choice has minimal practical impact on smoothing effectiveness, with both Gaussian and hyperbolic secant kernels providing equivalent benefits.

Examining the coefficient of variation (standard error relative to mean improvement) reveals that EST-PD maintains stable performance across bootstrap replications. Datasets with strong EST-PD performance (CCPP, Stock, Forest) show standard errors of 0.15-0.30 percentage points on 12-16\% improvements, yielding coefficients of variation under 2.5\%. Even on challenging datasets (for example, Real Estate), where mean improvements are modest, standard errors remain proportionally small (0.42 on 3.0\% mean improvement), suggesting that the limited benefit is consistent rather than arising from high variability. As such, we recommend the EST-PD configuration as a reasonable default choice for our smoothed random forests.
}

\section{Conclusion}

We introduced a novel approach to smoothing piecewise constant functions, such as decision trees and random forests. The key idea is to convolve the output of the tree with a smoothing kernel; this can be done via a latent variable representation, in which the randomness in the latent variable can be shown to mimic the variability in estimating change points in a tree under resampling. Experimental results demonstrate that this post-hoc smoothing step can substantially improve standard random forests, in terms of predictive accuracy, while remaining computationally tractable. 

This work has focused on estimating the conditional mean of the regression function, but the idea of post-hoc smoothing in random forests could also be extended to other statistics, such as the conditional variance (for uncertainty quantification). Another important direction is extending the framework to classification tasks by applying kernel smoothing to log-odds predictions for binary classification or logit-transformed class probabilities for multi-class problems, which would preserve probabilistic properties while introducing smoothness in decision boundaries. These extensions represent topics of ongoing research. 

While the framework developed here focuses on random forests, the approach could theoretically be extended to encompass gradient boosting methods such as XGBoost and LightGBM, which also produce piecewise constant predictions. However, unlike random forests, where trees are built independently, gradient boosting constructs trees sequentially, with each correcting previous errors. Initial experiments suggest that this sequential dependency means that the global calibration approach used here cannot be directly applied. Instead, per-tree calibration appears to be necessary, which substantially increases computational complexity from optimising 2 global parameters to $2T$ tree-specific parameters for a $T$-tree ensemble. However, given the performance improvements that are possible via tree smoothing, extending this methodology to boosted tree ensembles represents an important direction for future work.

\backmatter

\bmhead{Acknowledgements}

This work was supported by the Australian Research Council (DP210100045).


\begin{appendices}

\section{Proofs}\label{secA1}
\newrevtext{
\subsection{Differentiability of the Smoothed Prediction Function}
\label{sec:differentiability:smoothed}

In contrast to standard decision trees, smoothing the prediction function of a decision tree yields a surface that is differentiable, provided that the kernel used is itself differentiable. 
%
%
The derivative of the prediction function \eqref{piece_prediction} with respect to the $j$-th input coordinate of \( \bx_0 \), is given by an integral involving the product of kernel evaluations across all dimensions other than $j$, and the derivative of the kernel in dimension $j$:
%
%
%
\begin{equation*}
\label{multi_der}
    \begin{split}
    &\frac{\partial\hat{{y}}(\bx_0\given \bX,\by,\boldlambda)}{\partial\bx_0^j} = \\
        &\beta_1\sum_{i=1}^k  c_i\left[\left(\prod_{d=1,d\neq j}^p \intop_{D_i^d}k(\bz^d\given \bx_0^d,\lambda_d) \diff \bz^d\right)\left(\intop_{D_i^j} k'(\bz^j\given \bx_0^j,\lambda_j) \diff \bz^j\right)\right]
    \end{split}
\end{equation*}
where $k'(\bz^j \given \bx_0^j, \lambda_j) := \frac{\partial}{\partial \bx_0^j} 
k(\bz^j \given \bx_0^j, \lambda_j)$. These derivatives take the form
\begin{equation*}
\label{k_prime_gaussian}
    k'(z \given x_0, \lambda) = \left(\frac{z - x_0}{\lambda^2}\right) k(z \given x_0, \lambda)
\end{equation*}
for the Gaussian kernel, and
\begin{equation*}
\label{k_prime_hyper}
    k'(z \given x_0, \lambda) = 
    \left(\frac{\pi}{4\lambda^2}\right)
    \operatorname{sech}\!\left(\frac{\pi(z-x_0)}{2\lambda}\right)
    \tanh\!\left(\frac{\pi(z-x_0)}{2\lambda}\right)
\end{equation*}
for the hyperbolic secant kernel, respectively.
}

\newrevtext{
\subsection{Shrinkage Property of Probabilistic Smoothing}
\label{sec:basic_properties}
For $f \in L_2$ and symmetric kernels, i.e., $k(\bz; \bx, \boldlambda)=k(\bx; \bz, \boldlambda)$, we have 
\begin{align*}
\|\tilde{f}\|_2^2 &= \int_{\R^p} \left(\int_{\R^p} f(\bz)k(\bz; \bx, \boldlambda) \mathrm{d}\bz\right)^2 \mathrm{d}\bx \\
&\leq \int_{\R^p} \int_{\R^p} f(\bz)^2k(\bz; \bx, \boldlambda) \mathrm{d}\bz \, \mathrm{d}\bx \\
&= \int_{\R^p} f(\bz)^2 \int_{\R^p} k(\bz; \bx, \boldlambda)   \mathrm{d}\bx\, \mathrm{d}\bz \\
&= \int_{\R^p} f(\bz)^2 \underbrace{\int_{\R^p} k(\bx; \bz, \boldlambda)  \mathrm{d}\bx}_{=1} \, \mathrm{d}\bz = \|f\|_2^2 \enspace .
\end{align*}
Here, we used Jensen's inequality for the inequality step and the symmetry of normalisation of $k$ in the last step.

Note that tree and forest models are usually not directly square-integrable. However, we can assume a bounded domain $\mathcal X \subseteq \R^p$ of interest and simply set $f(\bx) = 0$ for $\bx \not\in \mathcal{X}$. 
A piecewise constant function with bounded support is clearly in $L_2$, hence the above result establishes $L_2$-shrinkage with respect to such modified tree and forest prediction functions.}

\subsection{Proof of Theorem~\ref{theorem:1}}
\label{sec:proof_thm1}
The (conditional) distributions of design points on the left and right of the split are $x \given x<b \sim U(b-w,w)$ and $x \given x>c \sim U(b,b+w)$. Without loss of generality, assume that the design points $x_1,\ldots,x_n$ are sorted in ascending order, and let $\tilde{n} = {\sup}_I \{ i : y_i=0\}$ denote the number of design points on the left of the cutpoint. Write $x_{\sup} = x_{\tilde{n}}$ and $x_{\inf} = x_{\tilde{n}+1}$ for the two design points that sit either side of the cut-point. Let $p_n = \tilde{n}/n$; then, by Theorem~9.6 from \cite{van2000asymptotic} (page 129) we have
\[
    \frac{n p_n (x_{\inf} - b)}{w} \overset{d}{\to} -{\rm Exp}(1) \; \; {\rm and} \; \; \frac{n (1-p_n)(x_{\sup} - b)}{w} \overset{d}{\to} {\rm Exp}(1),
\]
where ${\rm Exp}(s)$ denotes an exponential distribution with scale $s$. We note that $p_n = 1/2 + o_p(1)$, so $n p_n = (1-p_n) n = n/2 + o_p(n)$; therefore 
%
    $\frac{n}{2} \left( x_{\inf}+x_{\sup} - 2b \right) \overset{d}{\to} {\rm La}(0, w)$
%
as $n \to \infty$, as the difference of two exponential distributions with scale $s$ follows a Laplace distribution with mean zero and scale $2s$.
\subsection{Proof of Proposition~\ref{prop:global_local_relationship}}
\label{sec:proof_linearity_smoothing}
Consider an ensemble model $\mathcal{M}$ composed of $n$ base models $\mathcal{M}_1,\mathcal{M}_2,\dots,\mathcal{M}_n$. For a given query point $\bx_0$, the prediction of the ensemble model is defined as a weighted linear combination of its constituent base models:
\begin{align*}
    f(\mathbf{x}_0\given \mathcal{M}) =\sum_{i=1}^n w_i f(\mathbf{x}_0\given{\mathcal{M}_i})\enspace ,
\end{align*}
where $w_i$ represents the weight assigned to the $i$-{th} base model. 
Therefore, the prediction of an ensemble constructed from individually smoothed base models is:
\begin{align*}
\sum_{i=1}^n w_i \tilde{f}(\bx_0\given\mathcal{M}_i,\boldlambda) = \sum_{i=1}^n w_i\int f(\bz\given\mathcal{M}_i)k(\bz|\mathbf{x}_0,\boldlambda)\diff\bz\enspace .
\end{align*} 
The smoothed prediction of the ensemble model $\mathcal{M}$ is:
\begin{align*}
\begin{split}
\tilde{f}(\bx_0\given\mathcal{M},\boldlambda) &= \E[f(\bz|\mathcal{M},\boldsymbol{\lambda})]\\
&=\int f(\bz\given\mathcal{M},\boldlambda)k(\bz\given\bx_0,\boldlambda)\diff\bz\\
& = \int \sum_{i=1}^n w_i f(\bz\given{\mathcal{M}_i})k(\bz\given\bx_0,\boldlambda)\diff\bz\\
& = \sum_{i=1}^n w_i \int  f(\bz\given{\mathcal{M}_i})k(\bz\given\bx_0,\boldlambda)\diff\bz\\
&= \sum_{i=1}^n w_i \tilde{f}(\bx_0\given\mathcal{M}_i,\boldlambda) \enspace .
\end{split}
\end{align*}
Therefore, under the assumption of a uniform smoothing parameter $\boldlambda$ applied across all ensemble members, the prediction obtained from a smoothed ensemble model is equivalent to the prediction obtained from an ensemble of individually smoothed models.

\revtext{
\section{Smoothing Parameter Optimisation}\label{lambda_opt}

The smoothing parameters $\boldlambda$ in our framework are optimised by minimising the out-of-bag (OOB) prediction error using gradient-based optimisation methods. We leverage PyTorch automatic differentiation to compute gradients of the OOB loss with respect to smoothing parameters, enabling efficient optimisation across all four smoothing modes.

\subsection{Differentiable OOB loss construction}
\label{sec:selection_details}
The smoothing parameters $\boldLambda$ are optimised by minimising the OOB prediction error as defined in Equation~\eqref{eq:oob_risk} for the four smoothing modes. The key enabler of gradient-based optimisation is that the smoothed predictions computed via Equation~\eqref{piece_prediction} are differentiable with respect to $\boldlambda$. Specifically, the kernel probability computation in Equation~\eqref{eq:prob:norm:kernel} involves products of standard CDF differences (normal or hyperbolic secant), which are smooth functions of the smoothing parameters. This structure allows us to compute $\nabla_{\boldlambda }\mathcal{L(\boldlambda)}$ efficiently using automatic differentiation in PyTorch, enabling gradient-based optimisation across all smoothing modes regardless of parameter dimensionality.

\subsection{Optimisation methods for different smoothing modes}

The choice of optimisation algorithm is tailored to the dimensionality of the parameter space, balancing convergence speed, stability, and computational efficiency.\\

\noindent\textbf{STE (smoothed tree ensemble)}. For the global smoothing parameter $\lambda \in \R_+$, we employ L-BFGS (limited-memory Broyden-Fletcher-Goldfarb-Shanno), a quasi-Newton method that approximates the Hessian using gradient history. L-BFGS is optimal for low-dimensional problems as it exploits second-order curvature information for superlinear convergence, requires minimal hyperparameter tuning (learning rate typically set to 1.0), and converges rapidly (typically 10-20 iterations). Configuration: L-BFGS with strong Wolfe line search, maximum 20 iterations per step, history size of 10 gradients.\\

\noindent\textbf{STE-PD (smoothed tree ensemble with per-dimension)}. For dimension-specific parameters $\boldlambda = (\lambda_1,\cdots,\lambda_p)\in \R_+^p$ where $p$ is typically 5-50, we use SGD with Nesterov momentum combined with cosine annealing learning rate scheduling. This configuration provides stable convergence through momentum (coefficient 0.9), efficient computation (first-order method, lower memory than L-BFGS), and smooth learning rate decay, avoiding sudden changes. Configuration: SGD with Nesterov momentum (momentum=0.9), initial learning rate 0.01, cosine annealing to minimum $10^{-6}$, weight decay $10^{-5}$.\\

\noindent\textbf{EST (ensemble of smoothed trees)}. For tree-specific parameters $\boldlambda = (\lambda_1,\cdots,\lambda_T) \in \R_+^T$, where $T$ is typically 50-200, we employ AdamW (Adam with decoupled weight decay) with ReduceLROnPlateau scheduling. AdamW is advantageous for moderately high-dimensional spaces because it maintains adaptive learning rates for each parameter independently, incorporates weight decay for regularisation (preventing overfitting with many parameters), and dynamically adjusts the learning rate based on OOB loss plateaus. Configuration: AdamW with initial learning rate 0.01, betas=(0.9, 0.999), weight decay $10^{-4}$. ReduceLROnPlateau reduces the learning rate by a factor of 0.5 after 10 epochs without improvement, with a minimum learning rate of $10^{-6}$.\\

\noindent\textbf{EST-PD (ensemble of smoothed trees with per-dimension)}: For the full parameter matrix $\Lambda = [\boldlambda_1,\cdots,\boldlambda_T]$ where $\boldlambda_t \in \R_+^p$, yielding $T\times p$ parameters (potentially hundreds to thousands), we use AdamW with OneCycleLR scheduling. This configuration addresses high-dimensional optimisation challenges through independent adaptive learning rates for each of the $T\times p$ parameters, OneCycleLR super-convergence strategy (warm-up phase at 30\% of training followed by cosine annealing), and strong regularisation (weight decay $10^{-3}$) to prevent overfitting. Configuration: AdamW with initial learning rate 0.01, betas=(0.9, 0.999), weight decay $10^{-3}$. OneCycleLR with a maximum learning rate of 0.1, 30\% warm-up, and cosine annealing.
}

\section{Statistical Analysis of Results}\label{sec:statistical_analysis}

\revtext{We conducted a non-parametric statistical evaluation to assess performance differences among methods using the {\tt autorank} package in Python. The analysis was based on 5,300 paired samples, representing 100 bootstrap replications across 53 dataset-training size combinations. The evaluation metric was percentage improvement in MSE relative to the RF(10) baseline, providing a scale-free comparison across datasets with widely differing MSE ranges.}\\
\begin{table}[ht]
\centering
\begin{tabular}{lrrllll}
\toprule
 & MR & MED & MAD & CI & $\gamma$ & Magnitude \\
\midrule
GP & 5.302 & -31.137 & 37.086 & [-34.277, -28.277] & 0.000 & negligible \\
RF(20) & 4.424 & 4.239 & 2.581 & [4.017, 4.487] & -0.908 & large \\
RF(50) & 3.420 & 6.631 & 3.071 & [6.343, 6.896] & -0.968 & large \\
RF(100) & 2.879 & 7.319 & 3.198 & [7.047, 7.598] & -0.985 & large \\
EST-PD(Hypsec) & 2.493 & 9.985 & 4.527 & [9.629, 10.461] & -1.050 & large \\
EST-PD(Norm) & 2.482 & 9.963 & 4.482 & [9.579, 10.409] & -1.049 & large \\
\bottomrule
\end{tabular}
\caption{\revtext{Statistical summary of method performance based on percentage improvement in MSE relative to RF(10). MR: mean rank across all experimental conditions (lower is better); MED: median percentage improvement; MAD: median absolute deviation; CI: 95\% confidence interval for median; $\gamma$: Cliff's Delta effect size relative to RF(10) baseline (negative values indicate improvement over baseline); Magnitude: effect size classification.}}
\label{tbl:stat_results}
\end{table}

\revtext{\noindent\textbf{Normality testing}. We first applied Shapiro-Wilk tests to assess the normality of distributions for each method. All populations rejected the normality assumption ($p<0.001$ for all methods: GP, RF(20), RF(50), EST-PD(Norm) and EST-PD(Hypsec)), necessitating the non-parametric methods for subsequent analysis.\\

\noindent\textbf{Omnibus test}. Given the non-normal distributions and the presence of more than two populations, we employed the Friedman test as an omnibus test to determine whether significant differences exist among the median values of the populations. The Friedman test strongly rejected the null hypothesis of no difference in central tendency ($p<0.001$), indicating statistically significant differences among the methods.}\\

\revtext{\noindent\textbf{Post-hoc pairwise comparisons}. We applied the Nemenyi post-hoc test to identify specific pairwise differences between methods. The critical distance (CD) for the Nemenyi test at the significance level $\alpha = 0.05$ was calculated as CD = 0.104. Differences between methods are considered statistically significant if the difference in mean ranks exceeds this critical distance.\\

\noindent\textbf{Effect size estimation}. We computed Cliff's Delta ($\gamma$) as a measure of effect size for pairwise comparisons against the baseline RF(10). Cliff Delta ranges from -1 to 1, with values near -1 indicating great improvement over the baseline. We classify effect sizes as negligible ($|\gamma|<0.147$), small ($0.147\leq |\gamma|<0.33$), medium ($0.33\leq |\gamma|<0.474$), or large ($|\gamma|\geq 0.474$).\\

Table~\ref{tbl:stat_results} presents the statistical summary for all methods, including mean rank (MR), median improvement (MED), median absolute deviation (MAD), 95\% confidence intervals (CI), effect sizes ($\gamma$), and effect magnitude classifications. EST-PD(Norm) achieves the best mean rank (2.482), followed closely by EST-PD(Hypsec) (2.493). Both EST-PD variants demonstrate large effect sizes ($\gamma \approx -1.05$) relative to the baseline. The Nemenyi test reveals that EST-PD(Norm) and EST-PD(Hypsec) form a statistical equivalence group, with all other pairwise differences being statistically significant at the $\alpha=0.05$ level.
}

\section{Detailed Experimental Results}\label{apdx:detaile_exp}

This section presents detailed experimental results across all datasets, training sizes, and smoothing strategies. We report detailed performance metrics, including mean squared error (MSE), percentage improvement in MSE (\PIMSE) relative to the RF(10) baseline, median absolute deviation (MAD), and maximum absolute error. Additionally, we provide complete results for all four smoothing parameter strategies (STE, STE-PD, EST, EST-PD) with both Gaussian and hyperbolic secant kernels, enabling comparison of alternative approaches beyond the EST-PD method featured in the main text.

Table~\ref{tab:mse_dataset_mean} reports mean MSE values for each method across different training sizes, providing the absolute prediction error for each experimental condition. Table~\ref{tab:pimse_dataset_mean} presents the corresponding percentage improvements in MSE relative to the RF(10) baseline, showing the relative performance gains achieved by each method.

\begin{table}[htbp]
\resizebox{\columnwidth}{!}{%
\begin{tabular}{rccrrrrrrr}
\hline
\multicolumn{1}{c}{Data} &
  P &
  n &
  \multicolumn{1}{c}{RF(10)} &
  \multicolumn{1}{c}{RF(20)} &
  \multicolumn{1}{c}{RF(50)} &
  \multicolumn{1}{c}{RF(100)} &
  \multicolumn{1}{c}{GP} &
  \multicolumn{1}{c}{EST-PD(Norm)} &
  \multicolumn{1}{c}{EST-PD(Hypsec)} \\ \hline
CCPP                  & 4  & 50  & 35.40 & 33.43 & 32.48 & 32.32 & 51.74  & 31.31 & 30.88 \\
CCPP                  & 4  & 100 & 28.32 & 26.87 & 25.97 & 25.80 & 46.85  & 24.64 & 24.44 \\
CCPP                  & 4  & 200 & 24.65 & 23.60 & 22.98 & 22.75 & 42.40  & 21.43 & 21.30 \\
CCPP                  & 4  & 300 & 23.48 & 22.50 & 21.95 & 21.78 & 43.45  & 20.48 & 20.41 \\
CCPP                  & 4  & 400 & 22.19 & 21.36 & 20.87 & 20.66 & 43.45  & 19.52 & 19.48 \\
CCPP                  & 4  & 500 & 21.26 & 20.39 & 19.95 & 19.77 & 40.44  & 18.77 & 18.74 \\
Qsar Fish Toxicity    & 6  & 50  & 1.32  & 1.27  & 1.25  & 1.24  & 1.51   & 1.22  & 1.22  \\
Qsar Fish Toxicity    & 6  & 100 & 1.16  & 1.12  & 1.10  & 1.09  & 1.41   & 1.06  & 1.06  \\
Qsar Fish Toxicity    & 6  & 200 & 1.06  & 1.02  & 1.00  & 0.99  & 1.31   & 0.97  & 0.97  \\
Qsar Fish Toxicity    & 6  & 300 & 1.00  & 0.97  & 0.94  & 0.94  & 1.20   & 0.92  & 0.92  \\
Qsar Fish Toxicity    & 6  & 400 & 0.98  & 0.94  & 0.92  & 0.92  & 1.16   & 0.90  & 0.90  \\
Qsar Fish Toxicity    & 6  & 500 & 0.94  & 0.91  & 0.88  & 0.88  & 1.11   & 0.87  & 0.87  \\
Real Estate           & 6  & 50  & 91.46 & 88.84 & 87.43 & 86.80 & 167.96 & 90.41 & 89.94 \\
Real Estate           & 6  & 100 & 82.24 & 78.47 & 76.03 & 75.21 & 157.59 & 77.27 & 76.94 \\
Real Estate           & 6  & 200 & 70.75 & 68.72 & 66.78 & 66.07 & 142.03 & 67.90 & 67.79 \\
Real Estate           & 6  & 300 & 64.29 & 61.11 & 59.35 & 59.18 & 127.78 & 60.56 & 60.40 \\
Real Estate           & 6  & 400 & 62.14 & 59.72 & 57.88 & 57.52 & 121.42 & 60.75 & 60.76 \\
Yacht Hydrodynamics   & 6  & 50  & 34.36 & 30.23 & 26.17 & 25.27 & 135.52 & 24.29 & 24.17 \\
Yacht Hydrodynamics   & 6  & 100 & 13.54 & 11.43 & 9.32  & 9.03  & 64.78  & 9.63  & 9.72  \\
Yacht Hydrodynamics   & 6  & 200 & 5.31  & 3.91  & 3.24  & 3.12  & 14.82  & 3.92  & 3.95  \\
Yacht Hydrodynamics   & 6  & 300 & 3.50  & 2.69  & 2.24  & 2.08  & 3.88   & 2.64  & 2.67  \\
Qsar Aquatic Toxicity & 8  & 50  & 2.24  & 2.15  & 2.13  & 2.12  & 2.35   & 2.09  & 2.09  \\
Qsar Aquatic Toxicity & 8  & 100 & 1.97  & 1.90  & 1.86  & 1.84  & 2.30   & 1.82  & 1.82  \\
Qsar Aquatic Toxicity & 8  & 200 & 1.76  & 1.69  & 1.64  & 1.63  & 2.55   & 1.62  & 1.62  \\
Qsar Aquatic Toxicity & 8  & 300 & 1.64  & 1.56  & 1.52  & 1.51  & 2.68   & 1.53  & 1.53  \\
Qsar Aquatic Toxicity & 8  & 400 & 1.59  & 1.52  & 1.48  & 1.47  & 2.70   & 1.50  & 1.50  \\
Qsar Aquatic Toxicity & 8  & 500 & 1.51  & 1.44  & 1.41  & 1.39  & 2.71   & 1.42  & 1.42  \\
Fertility             & 9  & 50  & 0.04  & 0.04  & 0.04  & 0.04  & 0.04   & 0.03  & 0.03  \\
Fertility             & 9  & 100 & 0.04  & 0.04  & 0.04  & 0.04  & 0.04   & 0.03  & 0.03  \\
Stock                 & 11 & 50  & 0.00  & 0.00  & 0.00  & 0.00  & 0.00   & 0.00  & 0.00  \\
Stock                 & 11 & 100 & 0.00  & 0.00  & 0.00  & 0.00  & 0.00   & 0.00  & 0.00  \\
Stock                 & 11 & 200 & 0.00  & 0.00  & 0.00  & 0.00  & 0.00   & 0.00  & 0.00  \\
Stock                 & 11 & 300 & 0.00  & 0.00  & 0.00  & 0.00  & 0.00   & 0.00  & 0.00  \\
Stock                 & 11 & 400 & 0.00  & 0.00  & 0.00  & 0.00  & 0.00   & 0.00  & 0.00  \\
Stock                 & 11 & 500 & 0.00  & 0.00  & 0.00  & 0.00  & 0.00   & 0.00  & 0.00  \\
Winequality (Red)     & 11 & 50  & 0.58  & 0.57  & 0.56  & 0.56  & 0.68   & 0.55  & 0.55  \\
Winequality (Red)     & 11 & 100 & 0.54  & 0.52  & 0.51  & 0.51  & 0.65   & 0.49  & 0.49  \\
Winequality (Red)     & 11 & 200 & 0.51  & 0.49  & 0.48  & 0.47  & 0.63   & 0.46  & 0.46  \\
Winequality (Red)     & 11 & 300 & 0.49  & 0.47  & 0.46  & 0.45  & 0.62   & 0.45  & 0.45  \\
Winequality (Red)     & 11 & 400 & 0.47  & 0.45  & 0.44  & 0.43  & 0.60   & 0.44  & 0.43  \\
Winequality (Red)     & 11 & 500 & 0.46  & 0.44  & 0.43  & 0.43  & 0.59   & 0.43  & 0.43  \\
Winequality (White)   & 11 & 50  & 0.75  & 0.73  & 0.71  & 0.71  & 0.82   & 0.70  & 0.70  \\
Winequality (White)   & 11 & 100 & 0.71  & 0.68  & 0.67  & 0.66  & 0.79   & 0.65  & 0.65  \\
Winequality (White)   & 11 & 200 & 0.66  & 0.63  & 0.62  & 0.61  & 0.77   & 0.60  & 0.61  \\
Winequality (White)   & 11 & 300 & 0.63  & 0.60  & 0.59  & 0.58  & 0.76   & 0.58  & 0.58  \\
Winequality (White)   & 11 & 400 & 0.61  & 0.59  & 0.57  & 0.57  & 0.76   & 0.57  & 0.57  \\
Winequality (White)   & 11 & 500 & 0.60  & 0.58  & 0.56  & 0.56  & 0.75   & 0.56  & 0.56  \\
Forest                & 12 & 50  & 2.56  & 2.51  & 2.47  & 2.46  & 2.10   & 2.17  & 2.18  \\
Forest                & 12 & 100 & 2.58  & 2.49  & 2.44  & 2.42  & 2.02   & 2.14  & 2.15  \\
Forest                & 12 & 200 & 2.48  & 2.41  & 2.37  & 2.35  & 1.99   & 2.09  & 2.09  \\
Forest                & 12 & 300 & 2.45  & 2.37  & 2.32  & 2.30  & 1.99   & 2.07  & 2.07  \\
Forest                & 12 & 400 & 2.40  & 2.33  & 2.28  & 2.26  & 1.95   & 2.03  & 2.03  \\
Forest                & 12 & 500 & 2.42  & 2.32  & 2.27  & 2.26  & 1.96   & 2.04  & 2.04  \\ \hline
\end{tabular}%
}
\caption{Mean MSE by dataset and training size. Each row reports the mean squared error averaged over 100 bootstrap replications for a specific dataset-training size combination. Lower values indicate better performance.}
\label{tab:mse_dataset_mean}
\end{table}
\begin{table}[htbp]
\resizebox{\columnwidth}{!}{%
\begin{tabular}{rcccccccc}
\hline
\multicolumn{1}{c}{Data} & P  & n   & RF(20) & RF(50) & RF(100) & GP      & EST-PD(Norm) & EST-PD(Hypsec) \\ \hline
CCPP                     & 4  & 50  & 5.26   & 8.06   & 8.42    & -48.31  & 12.02        & 13.21          \\
CCPP                     & 4  & 100 & 4.95   & 8.09   & 8.68    & -66.01  & 12.92        & 13.64          \\
CCPP                     & 4  & 200 & 4.18   & 6.66   & 7.61    & -72.13  & 13.00        & 13.52          \\
CCPP                     & 4  & 300 & 4.19   & 6.45   & 7.21    & -85.72  & 12.74        & 13.00          \\
CCPP                     & 4  & 400 & 3.73   & 5.93   & 6.84    & -96.63  & 11.97        & 12.11          \\
CCPP                     & 4  & 500 & 4.06   & 6.16   & 6.98    & -90.06  & 11.68        & 11.80          \\
Qsar Fish Toxicity       & 6  & 50  & 3.09   & 5.04   & 5.54    & -15.48  & 7.18         & 7.02           \\
Qsar Fish Toxicity       & 6  & 100 & 3.41   & 5.44   & 6.03    & -22.03  & 8.66         & 8.48           \\
Qsar Fish Toxicity       & 6  & 200 & 3.76   & 6.13   & 6.90    & -22.87  & 9.14         & 9.13           \\
Qsar Fish Toxicity       & 6  & 300 & 3.56   & 5.73   & 6.30    & -19.91  & 8.31         & 8.34           \\
Qsar Fish Toxicity       & 6  & 400 & 3.57   & 5.39   & 6.10    & -18.69  & 7.57         & 7.57           \\
Qsar Fish Toxicity       & 6  & 500 & 3.45   & 5.96   & 6.65    & -17.94  & 7.40         & 7.45           \\
Real Estate              & 6  & 50  & 2.59   & 3.79   & 4.51    & -91.42  & -0.03        & 0.49           \\
Real Estate              & 6  & 100 & 3.86   & 6.31   & 7.20    & -98.83  & 4.37         & 4.73           \\
Real Estate              & 6  & 200 & 2.93   & 5.82   & 6.60    & -104.30 & 3.54         & 3.69           \\
Real Estate              & 6  & 300 & 5.07   & 7.93   & 8.14    & -103.54 & 5.48         & 5.71           \\
Real Estate              & 6  & 400 & 3.98   & 7.02   & 7.68    & -105.74 & 1.60         & 1.49           \\
Yacht Hydrodynamics      & 6  & 50  & 10.73  & 22.95  & 27.02   & -401.53 & 30.74        & 30.91          \\
Yacht Hydrodynamics      & 6  & 100 & 12.33  & 27.86  & 29.75   & -494.94 & 29.06        & 28.19          \\
Yacht Hydrodynamics      & 6  & 200 & 24.01  & 36.51  & 38.80   & -213.61 & 25.24        & 24.42          \\
Yacht Hydrodynamics      & 6  & 300 & 18.09  & 32.43  & 37.32   & -28.15  & 21.35        & 20.67          \\
Qsar Aquatic Toxicity    & 8  & 50  & 3.72   & 4.89   & 4.95    & -6.25   & 6.56         & 6.44           \\
Qsar Aquatic Toxicity    & 8  & 100 & 3.59   & 5.45   & 6.36    & -17.51  & 7.67         & 7.52           \\
Qsar Aquatic Toxicity    & 8  & 200 & 3.88   & 6.64   & 7.22    & -45.78  & 7.82         & 7.70           \\
Qsar Aquatic Toxicity    & 8  & 300 & 5.07   & 7.09   & 7.75    & -64.48  & 6.65         & 6.67           \\
Qsar Aquatic Toxicity    & 8  & 400 & 4.70   & 6.71   & 7.50    & -70.85  & 5.72         & 5.76           \\
Qsar Aquatic Toxicity    & 8  & 500 & 4.67   & 6.90   & 7.84    & -80.09  & 5.86         & 5.80           \\
Fertility                & 9  & 50  & 1.37   & 2.65   & 3.07    & -0.71   & 11.83        & 11.77          \\
Fertility                & 9  & 100 & 0.46   & 2.12   & 3.30    & -1.78   & 14.06        & 14.39          \\
Stock                    & 11 & 50  & 11.71  & 14.28  & 14.51   & -215.58 & 14.80        & 14.67          \\
Stock                    & 11 & 100 & 8.95   & 11.83  & 12.54   & -343.25 & 18.94        & 18.83          \\
Stock                    & 11 & 200 & 8.66   & 11.26  & 12.36   & -437.71 & 17.45        & 17.40          \\
Stock                    & 11 & 300 & 7.56   & 10.11  & 10.95   & -487.85 & 15.70        & 15.66          \\
Stock                    & 11 & 400 & 7.42   & 10.33  & 11.32   & -511.57 & 15.47        & 15.33          \\
Stock                    & 11 & 500 & 6.59   & 9.37   & 10.46   & -541.41 & 14.99        & 14.93          \\
Winequality (Red)        & 11 & 50  & 2.21   & 2.99   & 3.63    & -17.89  & 5.50         & 5.62           \\
Winequality (Red)        & 11 & 100 & 3.49   & 5.46   & 5.89    & -21.67  & 8.07         & 8.07           \\
Winequality (Red)        & 11 & 200 & 3.93   & 5.98   & 6.78    & -25.33  & 8.76         & 8.80           \\
Winequality (Red)        & 11 & 300 & 4.13   & 6.41   & 7.24    & -26.11  & 7.92         & 7.84           \\
Winequality (Red)        & 11 & 400 & 4.11   & 6.74   & 7.51    & -28.34  & 7.46         & 7.54           \\
Winequality (Red)        & 11 & 500 & 4.65   & 7.08   & 8.01    & -28.63  & 7.24         & 7.35           \\
Winequality (White)      & 11 & 50  & 3.66   & 5.75   & 6.22    & -9.10   & 6.19         & 6.21           \\
Winequality (White)      & 11 & 100 & 3.66   & 6.01   & 6.71    & -12.10  & 7.78         & 7.73           \\
Winequality (White)      & 11 & 200 & 3.66   & 6.15   & 6.98    & -18.07  & 7.81         & 7.73           \\
Winequality (White)      & 11 & 300 & 4.26   & 6.71   & 7.63    & -21.13  & 7.43         & 7.35           \\
Winequality (White)      & 11 & 400 & 4.06   & 6.46   & 7.35    & -23.27  & 7.16         & 7.17           \\
Winequality (White)      & 11 & 500 & 4.60   & 6.86   & 7.78    & -23.43  & 7.05         & 6.97           \\
Forest                   & 12 & 50  & 1.89   & 3.35   & 3.81    & 17.18   & 14.81        & 14.54          \\
Forest                   & 12 & 100 & 3.20   & 5.18   & 6.04    & 21.05   & 16.56        & 16.36          \\
Forest                   & 12 & 200 & 2.82   & 4.37   & 4.98    & 19.53   & 15.68        & 15.55          \\
Forest                   & 12 & 300 & 3.03   & 5.01   & 5.77    & 18.52   & 15.49        & 15.38          \\
Forest                   & 12 & 400 & 2.86   & 4.94   & 5.71    & 18.78   & 15.49        & 15.38          \\
Forest                   & 12 & 500 & 3.95   & 6.04   & 6.77    & 18.84   & 15.52        & 15.50          \\ \hline
\end{tabular}%
}
\caption{Mean percentage improvement in MSE relative to RF(10) baseline by dataset and training size. Each row reports the percentage improvement averaged over 100 bootstrap replications for a specific dataset-training size combination. Positive values indicate improvement over the baseline, while negative values indicate worse performance.}
\label{tab:pimse_dataset_mean}
\end{table}
\revtext{
Table~\ref{tab:four_smoothing_modes} compares the performance of all four smoothing parameter strategies across the 10 datasets. Results are averaged across all training sizes ($n=50$ to 500) for each dataset, with standard errors reported in parentheses. This comparison demonstrates the relative effectiveness of different parameterisation approaches and explains the reason EST-PD was selected as the recommended method for the main experimental evaluation.}
\begin{table}[htbp]
\resizebox{\columnwidth}{!}{%
\begin{tabular}{rccccccccc}
\hline
\multicolumn{1}{c}{Data} & $p$
& \multicolumn{2}{c}{STE}
& \multicolumn{2}{c}{STE-PD}
& \multicolumn{2}{c}{EST}
& \multicolumn{2}{c}{EST-PD} \\
\cmidrule(lr){3-4}\cmidrule(lr){5-6}\cmidrule(lr){7-8}\cmidrule(lr){9-10}
& & Norm & Hypsec & Norm & Hypsec & Norm & Hypsec & Norm & Hypsec \\ \hline
CCPP                  & 4  & 12.98 (0.21) & \textbf{13.17 (0.20)} & 12.30 (0.19) & 12.89 (0.19) & 11.34 (0.16) & 12.83 (0.16) & 12.39 (0.15) & 12.88 (0.16) \\
Qsar Fish Toxicity    & 6  & 5.66 (0.23)  & 5.79 (0.22)  & 2.81 (0.31)  & $-0.16$ (0.36) & 6.30 (0.25) & 4.73 (0.28) & \textbf{8.04 (0.18)} & 8.00 (0.18) \\
Real Estate           & 6  & $-11.38$ (1.02) & $-13.14$ (1.06) & $-22.15$ (1.00) & $-23.86$ (1.01) & $-28.39$ (1.12) & $-28.18$ (1.11) & 2.99 (0.42) & \textbf{3.22 (0.42)} \\
Yacht Hydrodynamics   & 6  & 1.60 (1.24)  & 1.70 (1.18)  & 24.65 (0.96) & 25.15 (1.01) & $-185.57$ (9.83) & $-319.36$ (16.36) & \textbf{26.60 (0.90)} & 26.05 (0.91) \\
Qsar Aquatic Toxicity & 8  & 4.64 (0.27)  & 4.86 (0.27)  & 2.25 (0.35)  & 0.02 (0.39)  & 5.28 (0.29) & 4.29 (0.31) & \textbf{6.72 (0.20)} & 6.65 (0.20) \\
Fertility             & 9  & 9.59 (0.90)  & 9.61 (0.89)  & 3.66 (1.31)  & 3.18 (1.32)  & 5.98 (1.23) & 5.40 (1.26) & 12.94 (0.72) & \textbf{13.08 (0.73)} \\
Stock                 & 11 & $-409.94$ (5.05) & $-413.99$ (5.06) & $-410.50$ (4.99) & $-413.99$ (5.06) & $-369.06$ (4.30) & $-386.94$ (4.56) & \textbf{16.22 (0.30)} & 16.14 (0.30) \\
Winequality (Red)     & 11 & 1.27 (0.27)  & 1.47 (0.27)  & $-8.57$ (0.40) & $-11.23$ (0.42) & $-3.29$ (0.35) & $-5.16$ (0.37) & 7.49 (0.19) & \textbf{7.54 (0.19)} \\
Winequality (White)   & 11 & 0.29 (0.30)  & 0.64 (0.29)  & $-7.80$ (0.39) & $-9.10$ (0.40) & $-5.48$ (0.36) & $-6.45$ (0.37) & \textbf{7.23 (0.16)} & 7.19 (0.16) \\
Forest                & 12 & 17.94 (0.24) & 17.90 (0.24) & 17.85 (0.23) & \textbf{18.14 (0.24)} & 17.54 (0.23) & 17.70 (0.23) & 15.59 (0.21) & 15.45 (0.21) \\ \hline
\end{tabular}%
}
\caption{Mean percentage improvement in MSE relative to RF(10) baseline 
for all four smoothing parameter strategies, averaged across all training 
sizes for each dataset. Standard errors are reported in parentheses. 
Bold indicates the best performing method for each dataset.}
\label{tab:four_smoothing_modes}
\end{table}

Table~\ref{tab:MAD_mean} reports median absolute deviation (MAD) values, while Table~\ref{tab:MAD_PI} presents percentage improvements relative to the RF(10) baseline. MAD provides a robust measure of prediction error dispersion that is less sensitive to outliers than MSE.

Table~\ref{tab:ME_mean} reports Maximum Error (ME) values, while Table~\ref{tab:ME_PI} details percentage improvements relative to the RF(10) baseline. ME quantifies worst-case performance by capturing the most extreme deviation, providing a critical upper bound on prediction error.
\begin{table}[htbp]
\resizebox{\columnwidth}{!}{%
\begin{tabular}{rcrrrrrrr}
\hline
\multicolumn{1}{c}{Data} &
  P &
  \multicolumn{1}{c}{RF(10)} &
  \multicolumn{1}{c}{RF(20)} &
  \multicolumn{1}{c}{RF(50)} &
  \multicolumn{1}{c}{RF(100)} &
  \multicolumn{1}{c}{GP} &
  \multicolumn{1}{c}{EST-PD(Norm)} &
  \multicolumn{1}{c}{EST-PD(Hypsec)} \\ \hline
CCPP                  & 4  & 3.16 & 3.09 & 3.05 & 3.04 & 3.97 & \textbf{3.01} & \textbf{3.01} \\
Qsar Fish Toxicity    & 6  & 0.56 & 0.55 & 0.54 & 0.54 & 0.60 & \textbf{0.53} & \textbf{0.53} \\
Real Estate           & 6  & 3.99 & 3.89 & 3.85 & \textbf{3.81} & 6.99 & 4.12 & 4.10 \\
Yacht Hydrodynamics   & 6  & 0.58 & 0.54 & 0.50 & 0.49 & 1.43 & \textbf{0.47} & \textbf{0.47} \\
Qsar Aquatic Toxicity & 8  & 0.73 & 0.71 & \textbf{0.70} & \textbf{0.70} & 0.93 & 0.71 & 0.71 \\
Fertility             & 9  & 0.12 & 0.12 & \textbf{0.11} & \textbf{0.11} & 0.13 & \textbf{0.11} & \textbf{0.11} \\
Stock                 & 11 & \textbf{0.00} & \textbf{0.00} & \textbf{0.00} & \textbf{0.00} & 0.01 & \textbf{0.00} & \textbf{0.00} \\
Winequality (Red)     & 11 & 0.43 & 0.42 & \textbf{0.41} & \textbf{0.41} & 0.74 & 0.43 & 0.43 \\
Winequality (White)   & 11 & 0.52 & 0.50 & 0.50 & \textbf{0.49} & 0.86 & 0.52 & 0.52 \\
Forest                & 12 & 0.91 & 0.89 & 0.87 & 0.86 & \textbf{0.48} & 0.70 & 0.70 \\ \hline
\end{tabular}%
}
\caption{Mean median absolute deviation (MAD) of prediction errors by 
dataset, averaged across all training sizes. Bold indicates the best 
performing method for each dataset.}
\label{tab:MAD_mean}
\end{table}
\begin{table}[htbp]
\resizebox{\columnwidth}{!}{%
\begin{tabular}{rcrrrrrr}
\hline
\multicolumn{1}{c}{Data} &
  P &
  \multicolumn{1}{c}{RF(20)} &
  \multicolumn{1}{c}{RF(50)} &
  \multicolumn{1}{c}{RF(100)} &
  \multicolumn{1}{c}{GP} &
  \multicolumn{1}{c}{EST-PD(Norm)} &
  \multicolumn{1}{c}{EST-PD(Hypsec)} \\ \hline
CCPP                  & 4  & 2.23 (0.09)  & 3.54 (0.10)  & 3.97 (0.10)  & -26.10 (1.27)  & 4.68 (0.11)  & \textbf{4.87 (0.11)}  \\
Qsar Fish Toxicity    & 6  & 2.29 (0.17)  & 3.76 (0.18)  & 4.35 (0.19)  & -6.14 (0.43)   & \textbf{6.55 (0.19)}  & \textbf{6.55 (0.19)}  \\
Real Estate           & 6  & 2.28 (0.26)  & 3.38 (0.28)  & \textbf{4.25 (0.29)}  & -76.41 (1.14)  & -3.56 (0.42) & -3.16 (0.42) \\
Yacht Hydrodynamics   & 6  & 6.04 (1.22)  & 12.32 (1.18) & 14.86 (1.11) & -105.68 (7.56) & 15.40 (1.14) & \textbf{15.60 (1.14)} \\
Qsar Aquatic Toxicity & 8  & 2.26 (0.21)  & 3.64 (0.22)  & \textbf{3.91 (0.23)}  & -30.24 (0.91)  & 2.42 (0.28)  & 2.50 (0.28)  \\
Fertility             & 9  & -0.86 (0.89) & 0.71 (0.86)  & 0.61 (0.93)  & -11.37 (1.67)  & 6.68 (1.03)  & \textbf{6.79 (1.04)}  \\
Stock                 & 11 & 4.31 (0.22)  & 5.88 (0.24)  & 6.48 (0.24)  & -115.50 (1.16) & \textbf{12.08 (0.28)} & 12.03 (0.29) \\
Winequality (Red)     & 11 & 1.42 (0.25)  & 2.94 (0.25)  & \textbf{3.46 (0.24)}  & -74.02 (1.57)  & -0.76 (0.34) & -0.63 (0.34) \\
Winequality (White)   & 11 & 2.48 (0.18)  & 3.98 (0.17)  & \textbf{4.51 (0.17)}  & -66.68 (1.12)  & -0.05 (0.34) & -0.08 (0.33) \\
Forest                & 12 & 2.01 (0.22)  & 4.46 (0.26)  & 5.06 (0.27)  & \textbf{47.55 (0.78)}   & 22.50 (0.33) & 22.26 (0.34) \\ \hline
\end{tabular}%
}
\caption{Mean percentage improvement in median absolute deviation (MAD) 
of prediction errors relative to RF(10) baseline, averaged across all 
training sizes for each dataset. Standard errors are reported in 
parentheses. Bold indicates the best performing method for each dataset.}
\label{tab:MAD_PI}
\end{table}
\begin{table}[htbp]
\resizebox{\columnwidth}{!}{%
\begin{tabular}{rcrrrrrrr}
\hline
\multicolumn{1}{c}{Data} &
  P &
  \multicolumn{1}{c}{RF(10)} &
  \multicolumn{1}{c}{RF(20)} &
  \multicolumn{1}{c}{RF(50)} &
  \multicolumn{1}{c}{RF(100)} &
  \multicolumn{1}{c}{GP} &
  \multicolumn{1}{c}{EST-PD(Norm)} &
  \multicolumn{1}{c}{EST-PD(Hypsec)} \\ \hline
CCPP                  & 4  & 46.42 & 46.44 & 46.48 & 46.29 & \textbf{44.89} & 45.53 & 45.60 \\
Qsar Fish Toxicity    & 6  & 5.09  & 5.06  & 5.03  & 5.03  & 5.37  & \textbf{4.98}  & 4.99  \\
Real Estate           & 6  & 61.06 & 60.90 & 60.54 & 60.65 & 64.75 & 59.71 & \textbf{59.66} \\
Yacht Hydrodynamics   & 6  & 16.24 & 15.19 & 14.22 & \textbf{14.01} & 29.79 & 14.95 & 15.00 \\
Qsar Aquatic Toxicity & 8  & 5.49  & 5.39  & 5.37  & 5.36  & 5.26  & 5.19  & \textbf{5.18}  \\
Fertility             & 9  & 0.53  & 0.53  & 0.52  & 0.52  & 0.53  & \textbf{0.51}  & \textbf{0.51}  \\
Stock                 & 11 & 0.04  & \textbf{0.03}  & \textbf{0.03}  & \textbf{0.03}  & 0.06  & \textbf{0.03}  & \textbf{0.03}  \\
Winequality (Red)     & 11 & 2.92  & 2.86  & 2.84  & 2.83  & 2.84  & \textbf{2.73}  & \textbf{2.73}  \\
Winequality (White)   & 11 & 3.78  & 3.71  & 3.67  & 3.65  & \textbf{3.28}  & 3.46  & 3.46  \\
Forest                & 12 & 5.72  & 5.67  & 5.65  & 5.65  & 5.58  & \textbf{5.55}  & \textbf{5.55}  \\ \hline
\end{tabular}%
}
\caption{Mean maximum absolute error of prediction errors by dataset,
averaged across all training sizes. Bold indicates the best performing
method for each dataset.}
\label{tab:ME_mean}
\end{table}
\begin{table}[!t]
\resizebox{\columnwidth}{!}{%
\begin{tabular}{rcrrrrrr}
\hline
\multicolumn{1}{c}{Data} &
  P &
  \multicolumn{1}{c}{RF(20)} &
  \multicolumn{1}{c}{RF(50)} &
  \multicolumn{1}{c}{RF(100)} &
  \multicolumn{1}{c}{GP} &
  \multicolumn{1}{c}{EST-PD(Norm)} &
  \multicolumn{1}{c}{EST-PD(Hypsec)} \\ \hline
CCPP                  & 4  & -0.12 (0.11) & -0.22 (0.13) & 0.18 (0.13)  & \textbf{3.03 (0.44)}   & 1.80 (0.13) & 1.63 (0.13) \\
Qsar Fish Toxicity    & 6  & 0.40 (0.25)  & 0.76 (0.30)  & 0.81 (0.31)  & -6.78 (0.58)  & \textbf{1.78 (0.27)} & 1.73 (0.27) \\
Real Estate           & 6  & 0.21 (0.29)  & 1.25 (0.34)  & 1.02 (0.34)  & -7.59 (0.67)  & 2.44 (0.40) & \textbf{2.48 (0.41)} \\
Yacht Hydrodynamics   & 6  & 5.18 (0.99)  & 11.07 (1.03) & \textbf{12.33 (1.05)} & -92.55 (3.79) & 6.21 (0.86) & 5.76 (0.88) \\
Qsar Aquatic Toxicity & 8  & 1.58 (0.26)  & 2.04 (0.30)  & 2.16 (0.32)  & 1.31 (0.75)   & 5.22 (0.28) & \textbf{5.33 (0.28)} \\
Fertility             & 9  & 0.91 (0.47)  & 1.39 (0.57)  & 2.14 (0.59)  & -0.42 (1.16)  & 4.61 (0.76) & \textbf{4.62 (0.76)} \\
Stock                 & 11 & 3.80 (0.33)  & 4.18 (0.36)  & 4.67 (0.38)  & -88.82 (1.71) & 5.89 (0.35) & \textbf{6.09 (0.35)} \\
Winequality (Red)     & 11 & 1.77 (0.22)  & 2.36 (0.26)  & 2.74 (0.27)  & 1.92 (0.45)   & 5.88 (0.27) & \textbf{6.03 (0.27)} \\
Winequality (White)   & 11 & 1.49 (0.21)  & 2.60 (0.24)  & 3.19 (0.23)  & \textbf{12.51 (0.38)}  & 8.03 (0.22) & 8.08 (0.22) \\
Forest                & 12 & 0.74 (0.22)  & 0.98 (0.26)  & 1.04 (0.26)  & 2.01 (0.40)   & \textbf{2.65 (0.27)} & 2.62 (0.27) \\ \hline
\end{tabular}%
}
\caption{Mean percentage improvement in maximum absolute error of 
prediction errors relative to RF(10) baseline, averaged across all 
training sizes for each dataset. Standard errors are reported in 
parentheses. Bold indicates the best performing method for each dataset.}
\label{tab:ME_PI}
\end{table}

\clearpage

\end{appendices}


\bibliography{sn-bibliography}

\end{document}